\documentclass[11pt,a4paper]{article}
\usepackage[hyperref]{emnlp2020}
\usepackage{times}
\usepackage{float}
\usepackage{booktabs}
\usepackage{tabularx}
\usepackage{threeparttable}
\usepackage{longtable}
\usepackage{latexsym}
\usepackage{ragged2e}
\usepackage{multirow}
\usepackage[ruled]{algorithm2e}
\usepackage[fleqn]{amsmath}
\usepackage{enumitem}
\usepackage{graphicx}
\usepackage{subfig}
\usepackage{paracol}
\usepackage{comment}
\newcommand{\argmax}{\mathop{\mathrm{argmax}}}

\usepackage{microtype}

\aclfinalcopy 

\title{Majority Voting with Bidirectional Pre-translation For Bitext Retrieval}

\author{Alex Jones \\
  Dartmouth College \\
  \texttt{alexander.g.jones.23@dartmouth.edu} \\\And
  Derry Tanti Wijaya \\
  Boston University \\
  \texttt{wijaya@bu.edu} \\}

\date{}

\begin{document}
\maketitle
\begin{abstract}

Obtaining high-quality parallel corpora is of paramount importance for training NMT systems. However, as many language pairs lack adequate gold-standard training data, a popular approach has been to mine so-called "pseudo-parallel" sentences from paired documents in two languages. In this paper, we outline some problems with current methods, propose computationally economical solutions to those problems, and demonstrate success with novel methods on the Tatoeba similarity search benchmark and on a downstream task, namely NMT. We uncover the effect of resource-related factors (i.e. how much monolingual/bilingual data is available for a given language) on the optimal choice of bitext mining approach, and echo problems with the oft-used BUCC dataset that have been observed by others. We make the code and data used for our experiments publicly available.\footnote{\href{https://github.com/AlexJonesNLP/alt-bitexts}{https://github.com/AlexJonesNLP/alt-bitexts}}

\end{abstract}

\section{Introduction}

Mining so-called "pseudo-parallel" sentences from sets of similar documents in different languages ("comparable corpora") has gained popularity in recent years as a means of overcoming the dearth of parallel training data for many language pairs. With increasingly powerful computational resources and highly efficient tools such as FAISS \citep{faiss} at our disposal, the possibility of mining billions of pseudo-parallel bitexts for thousands of language pairs to the end of training a multilingual NMT system has been realized. In particular, \citet{fan2020englishcentric} perform global mining over billions of sentences in 100 languages, resulting in a massively multilingual NMT system containing supervised data for 2200 language pairs. \\
Despite these astounding breakthroughs in high-resource engineering, many questions remain to be answered about bitext mining from a research perspective, ones with particular relevance to the \textit{low}-resource engineering case, i.e. contexts with limited computational resources. While \citet{fan2020englishcentric} yield impressive results using hundreds of GPUs, aggressive computational optimization, and a global bitext mining procedure (i.e. searching the entire target corpus for a source sentence match), how these results transfer to the low-computational-resource case is not clear. Moreover, the effect of circumstantial (e.g. the resources available for a given language or language pair) or linguistic (e.g. typological) factors on bitext mining performance remains highly understudied. In fact, we argue that efforts to scale this task have outpaced efforts to rigorously document and understand the factors which determine its outcome. \\
In this paper, we highlight the problematic nature of using similarity score thresholding \citep{artetxe-schwenk-2019-laser, wikimatrix, ccmatrix, fan2020englishcentric} for mining both gold-standard and pseudo-parallel sentences, in the latter case focusing on document-level mining from the Wikipedia corpora in medium-low-resource languages (namely English-Kazakh and English-Gujarati). We propose a heuristic method involving pre-translation of source and/or target sentences before mining, and show that particular variations of this approach outperform various similarity thresholds for mining pseudo-parallel sentences, as well as gold-standard bitexts in certain cases. On the gold-standard mining task, we establish what are, to our knowledge, benchmarks on dozens of languages, and perform a comprehensive breakdown of results by language resource capacity, showing the optimal mining method to be partially dependent on this resource factor.

\section{Related Work}

Mining pseudo-parallel sentences from paired corpora for the purposes of training NMT systems is a decades-old problem, and dozens of solutions have been tried, ranging from statistical or heuristic-based approaches \citep{zhao-vogel-adaptive, resnik-smith-2003-web, munteanu-etal-2004-improved, fung-cheung-2004-multi, munteanu-marcu-2006-extracting} to similarity-based, rule-based, and hybrid approaches \citep{azpeitia-etal-2017-weighted, azpeitia-stacc, BOUAMOR18.8, Hangya2018Unsupervised, schwenk-2018-filtering, ramesh-sankaranarayanan-2018-neural, artetxe-schwenk-2019-margin, artetxe-schwenk-2019-laser, hangya-fraser-2019-unsupervised, wikimatrix, ccmatrix, wu-etal-2019-machine, keung2020unsupervised, tran-criss-2020, kvapilikova-etal-2020-unsupervised, feng2020labse, fan2020englishcentric}. Benchmarks to measure performance on this task include the BUCC\footnote{\textsc{B}uilding and \textsc{U}sing \textsc{C}omparable \textsc{C}orpora} '17/18 datasets \citep{zweigenbaum-etal-2017-overview, zweigenbaum:hal-01898360}, whose task involves spotting gold-standard bitexts within comparable corpora, and the Tatoeba dataset \citep{artetxe-schwenk-2019-laser}, whose task involves matching gold-standard pairs in truly parallel corpora. \\
Relevant to similarity-based mining methods are well-aligned cross-lingual word and sentence embeddings, which are some of the oldest constructs in NLP and have been tackled using hundreds of diverse approaches. Even among relatively recent efforts, these approaches range from static, monolingual embeddings \citep{pennington-etal-2014-glove, mikolov-etal-dist, simple-baseline, skip-thought} to static, multilingual ones \citep{klementiev-etal-2012-inducing, ammar-2016, schwenk-douze-2017-learning} to contextualized, monolingual ones \citep{peters-etal-2018-deep, subramanian2018learning, devlin-etal-2019-bert, liu2020roberta, infersent, reimers-gurevych-2019-sentence} to contextualized, multilingual ones \citep{song-2019-mass2, NEURIPS2019_c04c19c2, conneau-etal-2020-unsupervised, reimers-gurevych-2020-making, feng2020labse, Wang*2020Cross-lingual}, including efforts at cross-lingual alignment \citep{xu-etal-2018-unsupervised, artetxe-etal-2018-robust2, schuster-etal-2019-cross, zhang-etal-2019-girls, Cao2020Multilingual}. In this paper, our approach centers around using contextualized, multilingual sentence embeddings for the task of bitext mining, although we mention attempts at rule-and-similarity-based hybrid methods in Section \ref{bucc} in the appendix.

For low resource languages where parallel training data is little to none, unsupervised NMT can play a crucial role \cite{artetxe2018emnlp, artetxe2019acl-umt, artetxe2019acl-bli, artetxe2018unsupervised, hoang-etal-2018-iterative, lample-2017-unmt, lample2018word,  lample-etal-2018-phrase, pourdamghani-etal-2019-translating, wu-etal-2019-machine}. However, previous works have only focused on high-resource and/or similar-to-English languages. Most recently, several works have questioned the universal usefulness of unsupervised NMT and showed its poor results for low-resource languages \cite{kim-etal-2020-unsupervised,marchisio-etal-2020-unsupervised}. They note the importance of linguistic similarity between source and target language, and domain proximity along with size and quality of the monolingual corpora, for good unsupervised NMT performance. They reason that since these conditions can hardly be satisfied in the case of low resource languages, they result in poor unsupervised performance for these languages. However, recently it has been shown that training a language model on monolingual data, followed by training with unsupervised MT objective and then training on mined comparable data \cite{unsupMTuseful2021} can improve MT performance for low resource languages. In this work, we explore the usefulness of our mined bitext using a similar pipeline. We show an improvement over using only supervised training data for low resource language MT. 


\section{Model selection}

\subsection{Cross-lingual Sentence Embeddings}

We initially experiment with XLM-RoBERTa \citep{conneau-etal-2020-unsupervised} for our bitext mining task, using averaged token embeddings or the [CLS] (final) token embedding as makeshift sentence embeddings. However, we replicate \citet{reimers-gurevych-2020-making}'s results in showing these ad-hoc sentence embeddings to have relatively poor performance on the BUCC '17/18 EN-FR train data \citep{zweigenbaum-etal-2017-overview, zweigenbaum:hal-01898360} compared to bona fide sentence embeddings like LASER \citep{artetxe-schwenk-2019-laser} and LaBSE \citep{feng2020labse}. Thus, we opt to use LaBSE as our sentence embedding model, using its implementation in the Sentence Transformers \footnote{\href{https://www.sbert.net}{https://www.sbert.net}} library, as LaBSE performs state-of-the-art (SOTA) or near-SOTA on the BUCC and Tatoeba datasets \citep{artetxe-schwenk-2019-laser}\footnote{\href{https://github.com/facebookresearch/LASER/tree/master/data/tatoeba/v1}{https://github.com/facebookresearch/LASER/tree/master/\\data/tatoeba/v1}}. Moreover, being more recent than LASER, LaBSE has been investigated less thoroughly in the context of this task.




\section{Methods} \label{Methods}

\subsection{Margin-based Mining}

For our primary mining procedure, we use margin-based mining as described in \citet{artetxe-schwenk-2019-margin}. Seeking to mitigate the hubness problem \citep{hubness}, margin scoring poses an alternative to raw cosine similarity in that it selects the candidate embedding that "stands out" the most from its $k$ nearest neighbors. We use the \textit{ratio} margin score, as described in \citet{artetxe-schwenk-2019-margin} and defined below: \\


\setlength{\mathindent}{-1cm}

\small
\begin{flalign*}
& (1) \\
& \text{score}(x,y) = \\\nonumber & \frac{\text{cos}(x,y)}{\frac{1}{2k}(\sum_{z\in NN_{k}(x)}\text{cos}(x,z) + \sum_{z\in NN_{k}(y)}\text{cos}(y,z))}
\end{flalign*} \label{margin-eq}

\normalsize

As in \citet{artetxe-schwenk-2019-margin}, we use $k=4$ for all our mining procedures. We acknowledge that $k$ is indeed a tuneable and important hyperparameter of KNN search, and that higher values of $k$ may work better for bitext mining in certain scenarios, depending on factors such as the size of the search space \citep{ccmatrix}. However, we don't make this hyperparameter a focus of this paper, instead addressing the problem of margin score thresholding and its relation to the size of the search space. We leave a thorough examination of $k$ and its effect on bitext mining performance for future work. \\
Additionally, \citet{artetxe-schwenk-2019-margin} describes four different "retrieval" techniques used to obtain sentence pairs after performing margin scoring, namely \textit{forward}, \textit{backward}, \textit{intersection}, and \textit{max score}. In the \textit{forward} procedure, every sentence in the source corpus is matched with some sentence in the target corpus, with this mapping being possibly non-surjective (i.e. not every sentence in the codomain need be mapped to). The \textit{backward} procedure is defined analogously, and the \textit{intersection} method (\textsc{INTERSECT} in Algorithm \ref{alg1}) takes the intersection of the resulting sentence pairs from these two procedures. Following \citet{artetxe-schwenk-2019-margin}, we find that \textit{intersection} produces good results, and use it on all mining tasks. \textit{Max score} takes the argmax sentence pair for any inconsistent alignments after bidirectional search (i.e. if sentences $x_j, y_k$ are paired in forward search and $x_l, y_k$ are paired in backward search, then take whichever has a higher associated margin score). Because \textit{max score} yields little or no benefit over \textit{intersection}, as shown in \citet{artetxe-schwenk-2019-margin}, we decided not to use it. We also try taking the union (denoted \textsc{UNION} in Algorithm \ref{alg1}) of forward and backward searches to prioritize recall, but find that this harms overall F1 on the BUCC '17/18 EN-FR training set due to decreased precision, and abandon the technique in further experiments. \\
We also perform all mining at the document level for the sake of computational thrift, and because recent approaches have targeted the global-level mining scenario but not verified the generalizablility of the techniques used. The Primary mining procedures described above are also outlined in Algorithm \ref{alg1}.

\LinesNumbered

\begin{algorithm}[t]
\small
\SetKw{Given}{Given}
\Given{$\mathcal{X}$, $\mathcal{Y}$, $k$, $t$ JOIN\_METHOD} \\
$\mathcal{X}$: Set of sentences in language X. May be grouped into documents or standalone sentences. \\
$\mathcal{Y}$: Set of parallel or comparable sentences in language Y. \\
$k$: Number of neighbors \\
JOIN\_METHOD: Method of combining sentence pairs after mining in the forward and backward directions. One of either INTERSECT or UNION. \\
$t$: Margin similarity threshold \\

\vspace{1em}


\newcommand\mycommfont[1]{\footnotesize\ttfamily\textcolor{blue}{#1}}
\SetCommentSty{mycommfont}

\footnotesize \textsc{Mine sentence pairs in both directions} \\
\small
\For{document $\mathcal{D}\in \mathcal{X}$}{
\For{$x\in \mathcal{D}$}{
                $nn_x \leftarrow NN(x, \mathcal{Y_D}, k)$ \tcp*{FAISS k-nearest neighbor search}
                $best_y = \argmax_{y\in nn_x} score(x, y)$ \tcp*{Eq.(1)}
                \small
                \If{$score(x, best_y) > t$}{
                $fwd_D \leftarrow (x, best_y)$
                } $fwd \leftarrow fwd_D$}}
\For{$\mathcal{D}\in \mathcal{Y}$}{\For{$y\in \mathcal{D}$}{$nn_y \leftarrow NN(y, \mathcal{X_D}, k)$
                $best_x = \argmax_{x\in nn_y} score(y, x)$ \\
                \If{$score(best_x, y) > t$}{
                $bwd_D \leftarrow (best_x, y)$
                } {$bwd \leftarrow bwd_D$}}}
\If{INTERSECT}{$\mathcal{P} \leftarrow \{fwd\}\cap\{bwd\}$}
\ElseIf{UNION}{$\mathcal{P} \leftarrow \{fwd\}\cup\{bwd\}$}

\Return{$\mathcal{P}$}

\caption{\small Doc-level margin-based mining} \label{alg1}
\end{algorithm}

\begin{algorithm}[h]
\small
\SetKw{Given}{Given}
\SetKw{IfNot}{if not}
\SetKw{Then}{then}

\Given{$\mathcal{X}, \mathcal{Y}, k, \boldsymbol{\mathcal{M}}, JOIN\_METHOD$} \\

$t$: \hspace{6pt} Margin score threshold \\
$\boldsymbol{\mathcal{M}}$: An NMT model \\

\If{TRANSLATE}{
                \If{EN\_TO\_XX}{
                                \For{$x\in \mathcal{X}$}{$\mathcal{X}_{trans} \leftarrow \boldsymbol{\mathcal{M}}(x \rightarrow lang_y)$
                                
                            $\mathcal{P}_{en\_xx} \leftarrow$
                            \tiny $\boldsymbol{AlgorithmI}(\mathcal{X}_{trans}, \mathcal{Y}, k, JOIN\_METHOD, t)$}
                            
                            \IfNot{\scriptsize \textsc{STRICT\_INT} or \textsc{PAIRWISE\_INT}} \Then
                            \hspace{15pt} {\Return{$\mathcal{P}_{en\_xx}$}}}
                            
                \If{XX\_TO\_EN}{
                                \For{$y\in \mathcal{Y}$}{$\mathcal{Y}_{trans} \leftarrow \boldsymbol{\mathcal{M}}(y \rightarrow lang_x)$
                                
                            $\mathcal{P}_{xx\_en} \leftarrow$
                            \tiny $\boldsymbol{AlgorithmI}(\mathcal{Y}_{trans}, \mathcal{X}, k, JOIN\_METHOD, t)$}
                            
                            \IfNot{\scriptsize \textsc{STRICT\_INT} or \textsc{PAIRWISE\_INT}} \Then
                            \\ \hspace{15pt} {\Return{$\mathcal{P}_{xx\_en}$}}
                
}}
$\mathcal{P}_{orig} \leftarrow$
\scriptsize
$\boldsymbol{AlgorithmI}(\mathcal{X}, \mathcal{Y}, k, JOIN\_METHOD, t)$
\small

\If{STRICT\_INT}{\Return{$\mathcal{P}_{orig}\cap\mathcal{P}_{en\_xx}\cap\mathcal{P}_{xx\_en}$}}

\ElseIf{PAIRWISE\_INT}{\Return{$\mathcal{P}_{orig}\cap\mathcal{P}_{en\_xx}\bigcup\mathcal{P}_{orig}\cap\mathcal{P}_{xx\_en}\bigcup\mathcal{P}_{en\_xx}\cap\mathcal{P}_{xx\_en}$}}

\Else{\Return $\mathcal{P}_{orig}$}

\caption{\small Secondary retrieval procedures} \label{alg2}
\end{algorithm}

\normalsize

\subsection{Filtering Procedures}
\subsubsection{Thresholding}

The most straightforward measure for filtering mined sentence pairs is setting a similarity score threshold, as shown in \citet{artetxe-schwenk-2019-margin}. Of course, there is a precision-recall tradeoff inherent to adjusting this threshold, and we show it is problematic in other ways for our document-level approach on a noisy corpus as well. We argue that choosing this threshold is an expensive and ambiguous process, one which has not been addressed with much rigor or been show to generalize to diverse mining scenarios.

\subsubsection{Pre-Translation}

Our approach capitalizes on multiple similarity-related signals by first translating either the source texts (i.e. en$\rightarrow$xx), target texts (xx$\rightarrow$en), or both. In our experiments on the Tatoeba dataset \citep{artetxe-schwenk-2019-laser}, we translate with Google Translate / GNMT \citep{gnmt} using Cloud Translation API. We also experimented with using \citet{tiedemann-thottingal-2020-opus}, but observed poor performance (e.g. poor coverage) for multiple language pairs. Due to the cost of using this API on large bodies of text, when mining on the English-Kazakh and English-Gujarati comparable corpora, we train a supervised system on WMT'19 data \citep{wmt2019}, with training corpora sizes given in Table \ref{tab:training-sizes}. When translating in either direction, we translate the entire corpus, e.g. translating all English sentences in the Wikipedia corpus to Kazakh.


\subsubsection{Strict \& Pairwise Intersection}

We also experiment with combining sentence pairs after mining using all three procedures described above and in Algorithm \ref{alg2}. We first mine using three approaches:
\begin{enumerate}[nosep]
    \item Mine sentence pairs using margin-based scoring (Algorithm \ref{alg1}) with the original en, xx sentences
    \item Mine pairs with the original en and translated xx$\rightarrow$en sentences
    \item Mine pairs with the original xx and translated en$\rightarrow$xx sentences
\end{enumerate}
After doing so, we either perform a "strict intersection" (\textsc{STRICT\_INT} in Algorithm \ref{alg2})—keeping only sentence pairs which appear in all three sets of pairs—or ``pairwise intersection" (\textsc{PAIRWISE\_INT}), a voting approach that keeps any pairs occurring in $\geq2$ of the sets above.


\subsection{Supervised and Unsupervised NMT}
We follow the same pipeline for training MT in \cite{unsupMTuseful2021} that is based on XLM \cite{conneau2019cross}. Following their pipeline, we first pretrain a bilingual Language Model (LM) using the Masked Language Model (MLM) objective \cite{devlin-etal-2019-bert} on the monolingual corpora of two languages (e.g. Kazakh and English for en-kk) obtained from Wikipedia, WMT 2018/2019\footnote{\href{http://data.statmt.org/news-crawl/}{http://data.statmt.org/news-crawl/}} and Leipzig corpora (2016)\footnote{\href{https://wortschatz.uni-leipzig.de/en/download/}{https://wortschatz.uni-leipzig.de/en/download/}}. For both the LM pretraining and NMT model fine-tuning, unless otherwise noted, we follow the hyper-parameter settings suggested in the XLM repository\footnote{ \href{http://github.com/facebookresearch/XLM}{http://github.com/facebookresearch/XLM}}. For every language pair we extract a shared 60,000 subword vocabulary using Byte-Pair Encoding (BPE) \cite{sennrich-etal-2016-neural}. After pretraining the LM, we train a NMT model in an unsupervised manner following the setup recommended in \citet{conneau2019cross}, where both encoder and decoder are initialized using the same pretrained encoder block. For training  unsupervised NMT, we use back-translation (\emph{BT}) and denoising auto-encoding (\emph{AE}) losses \cite{lample2018unsupervised}, and the same monolingual data as in LM pretraining. Lastly, to train a supervised MT using our mined comparable data, we follow \emph{BT+AE} with \emph{BT+MT}, where \emph{MT} stands for supervised machine translation objective for which we use the mined data. We stopped the training when the validation perplexity (LM pre-training) or BLEU (translation training) was not improved for ten checkpoints. We run all our experiments on 2 GPUs, each with 12GB memory. 

We compare the performance in terms of BLEU score of our MT model with a model that follows the same pipeline (LM pre-training, unsupervised MT training, followed by supervised MT training) but that uses (human translation) training data from WMT19 (Table~\ref{tab:training-sizes}). The size of the monolingual data we use for LM pretraining are also shown in  Table~\ref{tab:training-sizes}. 
\begin{table}[h]
    \centering
    \begin{tabularx}{0.485\textwidth}{p{9em}p{5em}p{5em}}
        \toprule
         \textbf{Train data} & \multicolumn{2}{l}{\textbf{Number of sentences}} \\ \hline
    
         & \textbf{en-kk} & \textbf{en-gu} \\
         \textbf{Monolingual} & 9.51M & 1.36M \\
         \textbf{Supervised} & & \\
         WMT'19 & 222,165 & 22,321 \\
         \hline
         \textbf{Comparable
         } & & \\
         \small %
         \textbf{1} LaBSE (threshold = 1.06) & \normalsize 430,762 & 120,989 \\
        \small
         \textbf{2} LaBSE (pairwise intersection, doc-level, all) & \normalsize 154,679 & 113,955 \\
         \small \textbf{3} LaBSE (pairwise intersection, doc-level, all, threshold = 1.20) & \normalsize 55,765 & — \\
         \small \textbf{4} LaBSE (pairwise intersection, doc-level, all, threshold = 1.35) & \normalsize 19,099 & — \\
        \bottomrule
    \end{tabularx}
    \caption{Sizes (in number of sentences) of training corpora used in training supervised and semi-supervised NMT. The comparable/pseudoparallel sentences are mined using margin-based scoring with LaBSE, with secondary retrieval procedures given in parentheses. These procedures are described in Section \ref{Methods}.}
    \label{tab:training-sizes}
\end{table}
\section{Experiments}


\subsection{Gold-standard Bitext Retrieval}

In gold-standard bitext retrieval tasks, the goal is to mine gold-standard bitexts from a set of parallel or comparable corpora. We use the common approach of finding k-nearest neighbors for each sentence pair (in both directions, if using \textsc{INTERSECT} in Algorithm \ref{alg1}), then choosing the sentence that maximizes the ratio margin score (Equation 1 in Section \ref{margin-eq}).


\paragraph{Tatoeba Dataset\footnote{\href{https://github.com/facebookresearch/LASER/tree/master/data/tatoeba/v1}{https://github.com/facebookresearch/LASER/tree/master/\\data/tatoeba/v1}}} \label{tatoeba-description}

The Tatoeba dataset, introduced by \citet{artetxe-schwenk-2019-laser}, contains up to 1,000 English-aligned, gold-standard sentence pairs for 112 languages. In light of our focus on lower-resource languages, we experiment only on the languages listed in Table 10 of \citet{reimers-gurevych-2020-making}, which are languages without parallel data for the distillation process they undertake. This heuristic choice is supported by relative performance against languages \textit{with} parallel data for distillation: the average raw cosine similarity baseline with LaBSE for the latter was 96.3, in contrast with 73.7 for the former. Specifically, the ISO 639-2 codes\footnote{\href{https://www.loc.gov/standards/iso639-2/php/code_list.php}{https://www.loc.gov/standards/iso639-2/php/code\_list.php}} for the languages we use are as follows:

\small afr, amh, ang, arq, arz, ast, awa, aze, bel, ben, ber, bos, bre, cbk, ceb, cha, cor, csb, cym, dsb, dtp, epo, eus, fao, fry, gla, gle, gsw, hsb, ido, ile, ina, isl, jav, ksb, kaz, khm, kur, kzj, lat, lfn, mal, mhr, nds, nno, nov, oci, orv, pam, pms, swg, swh, tam, tat, tel, tgl tuk, tzl, uig, uzb, war, wuu, xho, yid.

\normalsize


\paragraph{BUCC Dataset}

The BUCC '17/18 dataset \citep{zweigenbaum-etal-2017-overview, zweigenbaum:hal-01898360}, provided by the Workshop for Building and Using Comparable Corpora, features English-aligned comparable corpora from Wikipedia in French, German, Chinese, and Russian, with gold-standard bitexts from News Commentary inserted randomly throughout. The goal of the task is extract these gold-standard pairs, with performance measured using standard F1-score. This task has been tackled with a variety of approaches \citep{BOUAMOR18.8, etchegoyhen-azpeitia-2016-set, azpeitia-stacc, azpeitia-etal-2017-weighted, Hangya2018Unsupervised, artetxe-schwenk-2019-margin, artetxe-schwenk-2019-laser, hangya-fraser-2019-unsupervised, keung2020unsupervised, feng2020labse, reimers-gurevych-2020-making}. \\
We use only the publicly available\footnote{\href{https://comparable.limsi.fr/bucc2017/bucc2017-task.html}{https://comparable.limsi.fr/bucc2017/bucc2017-task.html}} EN-FR train data in our experiments, and initially experiment using rule-based metrics on top of margin-based mining, similar to \citet{keung2020unsupervised}. However, we note major problems with the BUCC data, which are discussed in Section \ref{bucc}, and for this reason—coupled with the lackluster performance of these rule-based metrics—do not report results on this dataset, though the methods we try are described in Section \ref{bucc}.


\subsection{Pseudo-parallel Sentences From Comparable Corpora}

In addition to gold-standard bitext mining, we also mine pseudo-parallel sentences from so-called comparable corpora. The aim of this task is as follows: given two sets of similar documents in different languages, find sentence pairs that are close enough to being translations to act as training data for an NMT system. Of course, unlike the gold-standard mining task, there are not ground-truth labels present for this task, and so evaluation must be performed on a downstream task like NMT.

    
\paragraph{Comparable Corpora}
Our comparable data is mined from comparable documents, which are linked Wikipedia pages in different languages obtained using the langlinks from Wikimedia dumps. For each sentence in a foreign language Wikipedia page, we use all sentences in its corresponding linked English language Wikipedia page as potential comparable sentences.

\paragraph{Pre-processing}

\begin{figure}[b]
    \includegraphics[width=0.35\textwidth]{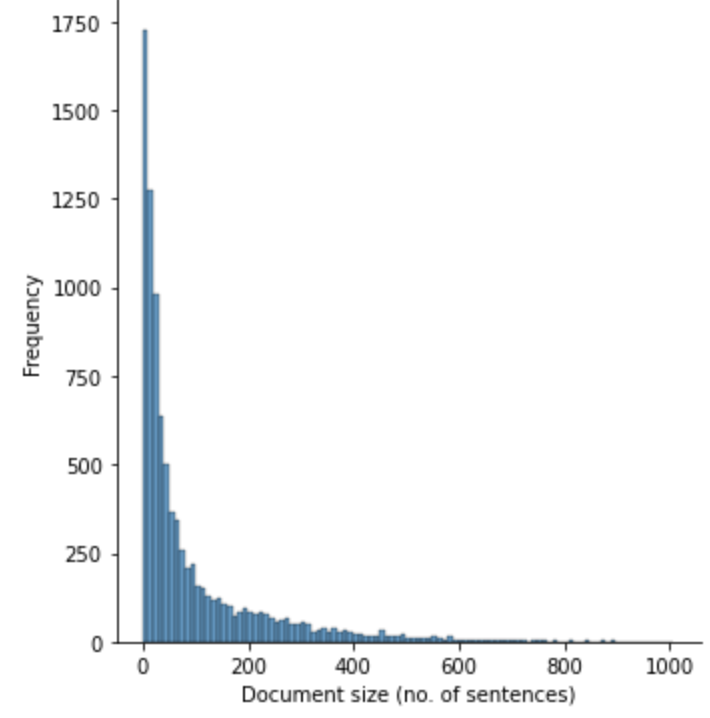}
    \caption{Distribution of document sizes in English component of EN-GU Wikipedia corpus.}
    \label{fig:doc-size}
\end{figure}

Since our comparable corpora for both EN-KK and EN-GU are grouped into documents, the most important pre-processing step we perform is eliminating especially short documents before similarity search. The motivation for this is that since we search at document-level, the quality of the resulting pairs could be highly degraded in particularly small search spaces, in a way that neither thresholding nor voting could mitigate. Note that average document length was much shorter for both Gujarati and Kazakh than for English, due simply to shorter Wikipedia articles in those languages. For the EN-KK corpus, we omitted any paired documents whose English version was $< 30$ words or whose Kazakh version was $< 8$ words, which we determined somewhat arbitrarily by seeing what values allowed for a sufficient number of remaining sentences. For the EN-GU corpus, we take a more disciplined approach and lop off the bottom $35\%$ of shortest document pairs, which happened to be $document\_length = 21$ sentences for English and $5$ sentences for Gujarati. This step accounted for the large number of documents in each corpus that contained very few sentences (see Figure \ref{fig:doc-size} for an example). \\
We performed additional more-or-less standard pre-processing, such as removing URLs, non-standard characters, and superfluous white space, as well as recurrent noise that we spotted in the corpora (such as ``href" in the English part of the EN-KK corpus).

    
\paragraph{Document-level mining vs. global mining}

Due to the sizes of the comparable corpora and our computational resources, we perform document-level mining (described in Algorithm \ref{alg1}) when retrieving pseudo-parallel sentence pairs for NMT training and global mining (mining over all sentence pairs in each corpus) when experimenting on the Tatoeba and BUCC corpora. Like \citet{ccmatrix, fan2020englishcentric}, we speculate that global mining yields better results than document-level mining, all else being equal. However, like \citet{wikimatrix}, we note that this conjecture has yet to be rigorously examined, and that we don't boast the resources to do so meaningfully.


\subsection{NMT}
We conduct experiments on Kazakh and Gujarati. They are spoken by 22M and 55M speakers worldwide, respectively, and are distant from English, in terms of writing scripts and alphabets. Additionally, these languages have few parallel but some comparable and/or monolingual data available, which makes them ideal and important candidates for our low-resource unsupervised NMT research. 

Our monolingual data for LM pre-training of these languages (shown in Table~\ref{tab:training-sizes}) are carefully chosen from the same topics (for Wikipedia) and the same domain (for news data). For the news data, we also select data from similar time periods (late 2010s) to mitigate domain discrepancy between source and target languages as per previous research \cite{kim-etal-2020-unsupervised}. We also downsample the English part of WMT NewsCrawl corpus so that our English and the corresponding foreign news data are equal in size. 

\section{Results \& Analysis}

\subsection{Tatoeba Dataset\footnote{\href{https://github.com/AlexJonesNLP/alt-bitexts/blob/main/source/retrieve_tatoeba_results.ipynb}{https://github.com/AlexJonesNLP/alt-bitexts/blob/main/source/retrieve\_tatoeba\_results.ipynb}}}

We mine bitexts on the Tatoeba test set in 64 different languages (listed in Section \ref{tatoeba-description}) using the primary mining procedure described in Algorithm \ref{alg1} with \textit{intersection} retrieval, in addition to seven different secondary mining procedures. The methods and corresponding results are reported in Table \ref{table_tatoeba_all} in terms of F1, and are summarized as follows, in the order in which they appear in the table:
\begin{enumerate}[nosep]
    \item Raw cosine similarity \citep{reimers-gurevych-2020-making}: find closest sentence pair using cosine similarity only
    \item "Vanilla" margin scoring: perform forward and backward searches and take intersection
    \item Margin scoring, threshold=1.06: margin scoring with a threshold of 1.06, à la \citet{ccmatrix} (Method \textbf{1} in Table \ref{tab:training-sizes})
    \item . . . threshold=1.20: optimal BUCC threshold
    \item Margin scoring using EN sentences translated to XX (Method 
    \item . . . using XX sentences translated to EN
    \item The strict intersection of pairs generated by methods 2, 5, and 6
    \item The pairwise intersection of pairs generated by method 2, 5, and 6 (Method \textbf{2})
\end{enumerate}
We report F1 instead of accuracy because the intersection methods (in both primary and secondary procedures) permit less than $100\%$ recall. \\
The results are broken down across languages by resource availability (as in "high-resource" or "low-resource"), as ranked on a 0-5 scale\footnote{\href{rb.gy/psmfnz}{rb.gy/psmfnz}}, and summarized in Table \ref{tatoeba-average}. Language-specific results are given in Table \ref{tatoeba-table-list}. \\
Because many of the languages in Table \ref{table_tatoeba_all} lack support in GNMT, the dominant method overall is vanilla margin scoring (Method 2 above), being the best-performing method on 28/64 languages\footnote{Note that 6/64 languages lack a resource categorization, so we report results on the remaining 58} and seeing an average gain over the baseline (Method 1) of $+5.2$ for all languages and $+6.9$ for languages on which it was the best-performing method. However, for languages with translation support, the pairwise intersection method (Method 8) won out, with an average gain over the baseline of $+4.0$, in contrast to vanilla margin scoring ($+3.6$). Moreover, pairwise intersection increased F1-score over vanilla margin scoring for 26/38 of these languages. In fact, among these 38 languages, vanilla margin scoring outperformed translation-based or hybrid (intersection) methods on only 11 languages, five of which were translated zero-shot (e.g. substituting Standard German for Low German or Esperanto for Ido when translating). \\
Simply translating non-English sentences into English before mining (Method 6) also performed well, netting best results on 18 languages and outperforming other methods on resource level 3 ($+4.3$ F1 over baseline) and level 4 ($+1.8$) languages. Meanwhile, pairwise intersection performed best on level 0 ($+7.3$) and level 2 ($+2.6$) languages, with vanilla margin scoring taking home the bread on level 1 ($+5.2$). Notably, thresholding (Methods 3\&4) almost exclusively did more harm than good (Method 3 achieved best results on only 3 languages, and Method 4 on none), and though reporting this may be viewed as a strawman attack on thresholding in the context of this task—identifying bitexts in gold-standard parallel corpora, as opposed to noisy comparable corpora—we note that gold-standard bitexts simply don't reliably lie beyond some set threshold, as shown in the right-two graphs in Figure \ref{fig:margin-plots}. Additionally, performing strict intersection (Method 7) led to decreased F1 due to dampened recall, suggesting majority voting is a better way to combine signals from similarity searches than all-or-nothing voting. We note as well that 6 languages on which vanilla margin scoring performs best are constructed (e.g. Esperanto, Ido) and 2 are extinct (Old English and Old Russian), inflating those results somewhat from a natural/living-language-focused perspective.

\subsection{NMT\footnote{\href{https://github.com/AlexJonesNLP/alt-bitexts/tree/main/source}{https://github.com/AlexJonesNLP/alt-bitexts/tree/main/source}}}

In Table \ref{tab:bleu-scores}, we show the performance in terms of BLEU scores of various NMT training schemes on the same WMT'19 test set. We train the supervised MT part of our pipeline system with gold-standard data (human translation WMT'19 data), our mined comparable/pseudoparallel ("silver-standard") data, and combinations of both i.e., training with comparable data followed by training with gold-standard data. We also provide Google Massively Multilingual MT performance on the same WMT'19 test set \cite{gnmt}. 

As we can see in Table \ref{tab:bleu-scores}, our method of mining bitext without thresholding (Method \textbf{2}) results in higher BLEU performance than bitext mined using margin scoring with a threshold of 1.06 (Method \textbf{1}), which is a commonly used threshold recommended by previous works for mining bitext using margin scoring. Method \textbf{2} also results in the best en$\rightarrow$gu performance, which outperforms previous unsupervised or supervised works. It outperforms the best previous work that uses WMT'19 data and iterative bitext mining by +3.3 BLEU. Since we do not perform iterative mining, if we consider the same previous work without iterative mining i.e., \citet{tran-criss-2020} Iter 1, ours outperforms that model by +12.1 BLEU in en$\rightarrow$gu direction and by +8.3 BLEU in gu$\rightarrow$en direction. 

When combined with supervised i.e., gold-standard data for training, our method for mining bitext which does not use any thresholding (Method \textbf{2}+WMT'19) also outperforms the same model which uses bitext mined using margin scoring with a threshold of 1.06 (Method \textbf{1}+WMT'19). Method \textbf{2}+WMT'19 also results in the best en$\rightarrow$kk performance, which outperforms previous unsupervised or supervised works. It outperforms the best previous work that uses WMT'19 data and iterative bitext mining by +4.7 BLEU. Since we do not perform iterative mining, if we consider the same previous work without iterative mining i.e., \citet{tran-criss-2020} Iter 1, ours outperforms that model by +5.6 BLEU in en$\rightarrow$kk direction and by +2.8 BLEU in kk$\rightarrow$en direction. It is also worth noting that for training our pipeline model we use fixed  hyper-parameter settings suggested in the XLM repository while previous works perform extensive hyperparameter tuning. We believe our performance can be improved further by tuning our hyperparameter settings. 

These results on low resource MT further demonstrate the superiority of our method for mining bitext without thresholding compared to margin scoring with thresholding for downstream low resource MT applications. 

\RaggedRight
\begin{table}[t!]
    \centering
    \small
    \begin{tabularx}{0.485\textwidth}{p{6em}cccc}
        \toprule
         \textbf{Corpus} & \multicolumn{4}{c}{\textbf{Language pair}} \\ \hline
         & \textbf{kk}$\rightarrow$\textbf{en} & \textbf{en}$\rightarrow$\textbf{kk} & \textbf{gu}$\rightarrow$\textbf{en} & \textbf{en}$\rightarrow$\textbf{gu} \\
          \textbf{Unsupervised} &&&& \\
         \citet{kim-etal-2020-unsupervised} & 2.0 & 0.8 & 0.6 & 0.6 \\
         \hline
         \textbf{Supervised} &&&& \\
         WMT'19 \citep{kim-etal-2020-unsupervised} & 10.3 & 2.4 & 9.9 & 3.5 \\
         WMT'19 \citep{tran-criss-2020} Iter 1 & \small 9.8 & 3.4 & 8.1 & 8.1 \\
         WMT'19 \citep{tran-criss-2020} Iter 3 & \small \textbf{13.2} & 4.3 & \textbf{18.0} & 16.9 \\ 
         Google MT \citep{gnmt} & \textbf{28.9} & \textbf{23.1} & \textbf{26.2} & \textbf{31.4} \\ 
        
        
        \hline
        
         \textbf{Our pipeline: unsup.+Sup.} &&&& \\
         WMT'19 & 11.2 & 7.3 & 5.7 & 10.2 \\
        Method \textbf{1} & 6.6 & 4.1 & 16.2 & 19.8 \\
        Method \textbf{2} & 8.6 & 6.1 & 16.4& \textbf{20.2 }\\ 
        Method \textbf{1}+WMT'19 
        & \small 11.8 & 7.9 & 15.4 & 18.5 \\
        Method \textbf{2}+WMT'19 
        & \small 12.6 & \textbf{9.0} & 15.8 & 19.1 \\
        
        \centerline{\rule{3cm}{0.4pt}}
        
        \textbf{\footnotesize Previous training procedure} &&&& \\
        Method \textbf{2}+WMT'19 
        & \small 11.8 & 7.9 & — & — \\
        Method \textbf{3}+WMT'19 & \small 11.8 & 8.1 & — & — \\
        Method \textbf{4}+WMT'19 & \small 12.2 & 8.5 & — & — \\
        LaBSE (threshold=1.20)+WMT'19 & \small 8.9 & 6.6 & — & — \\

    \bottomrule
    \end{tabularx}
    \caption{\footnotesize NMT training schemes and corresponding BLEU scores on WMT'19 test set. We train supervised systems with gold-standard data, comparable/pseudoparallel ("silver-standard") data, and combinations of both. We also try supplementing unsupervised training with each of these three types of supervised data, providing full supervision, weak supervision, or both. We also provide a benchmark from \citet{gnmt}. The methods listed in the table are given in Table \ref{tab:training-sizes}.}
    \label{tab:bleu-scores}
\end{table}

\justifying

\normalsize

\subsection{The Problem with Thresholding}

\begin{figure}[h!]
    \includegraphics[width=0.48\textwidth]{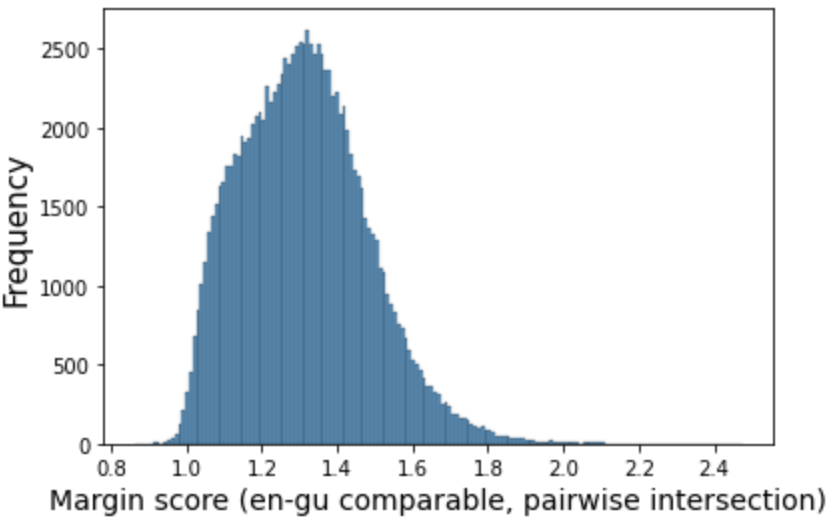}
    \caption{Margin scores on EN-GU pairs mined using the pairwise intersection method.}
    \label{fig:margin-mv}
\end{figure}

\begin{figure*}[t]
\centering
\subfloat[Margin scores on English-Kazakh comparable corpora]{\includegraphics[width=0.2\textwidth]{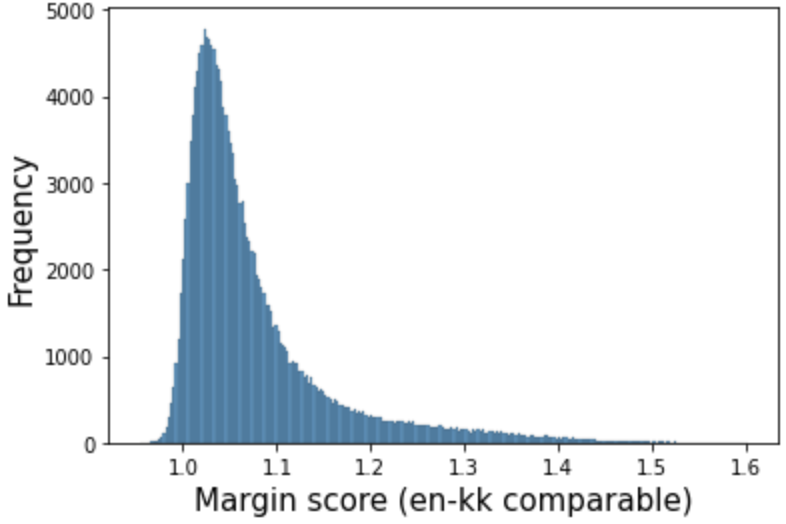}}
\subfloat[Margin scores on English-Gujarati comparable corpora]{\includegraphics[width=0.2\textwidth]{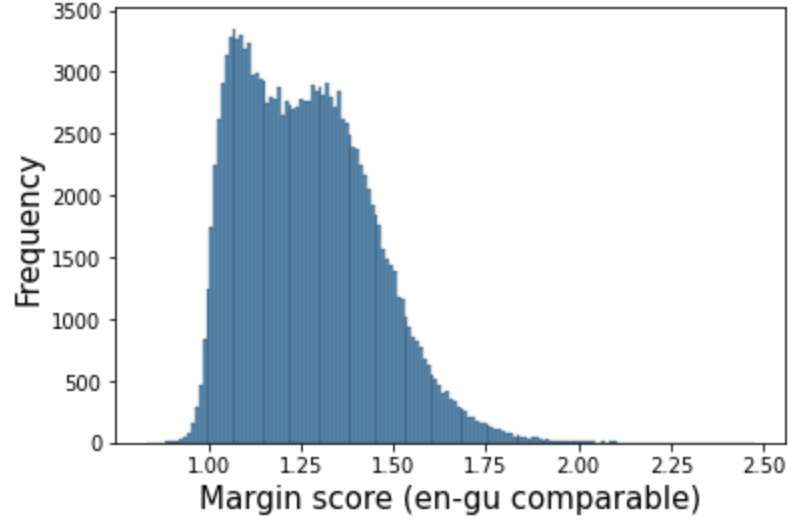}}
\subfloat[Margin scores on BUCC '17/18 English-French data]{\includegraphics[width=0.2\textwidth]{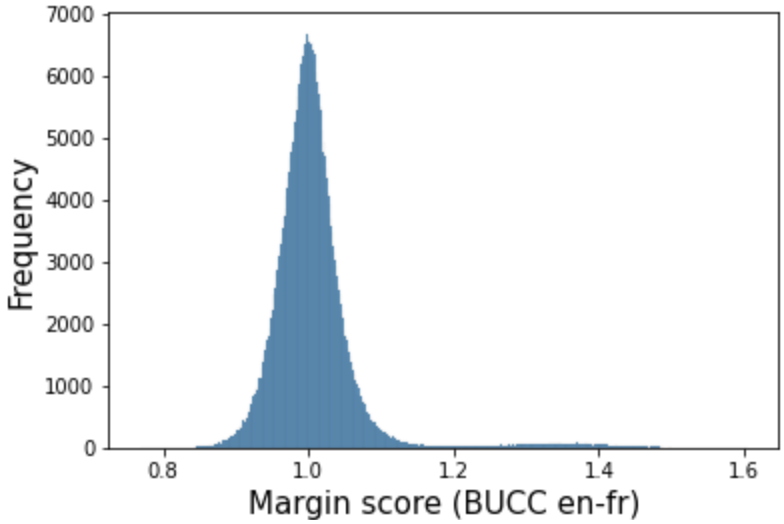}}
\subfloat[Margin scores on Tatoeba English-Maltese data]{\includegraphics[width=0.2\textwidth]{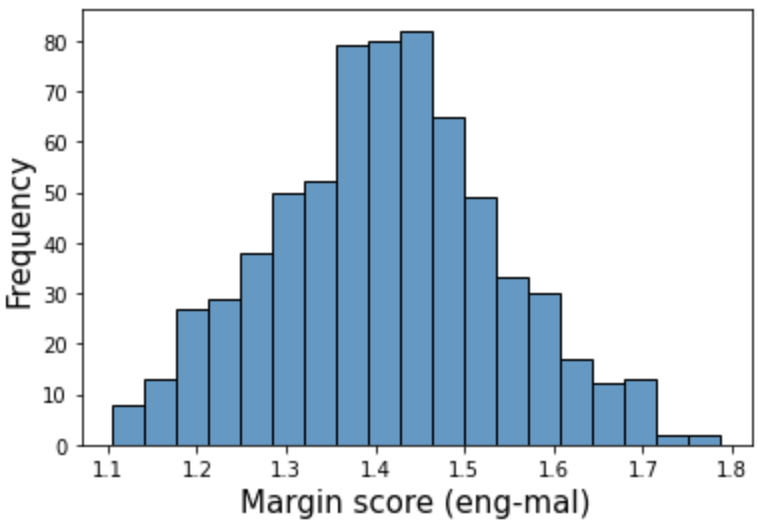}}
\subfloat[Margin scores on Tatoeba English-Telugu data]{\includegraphics[width=0.2\textwidth]{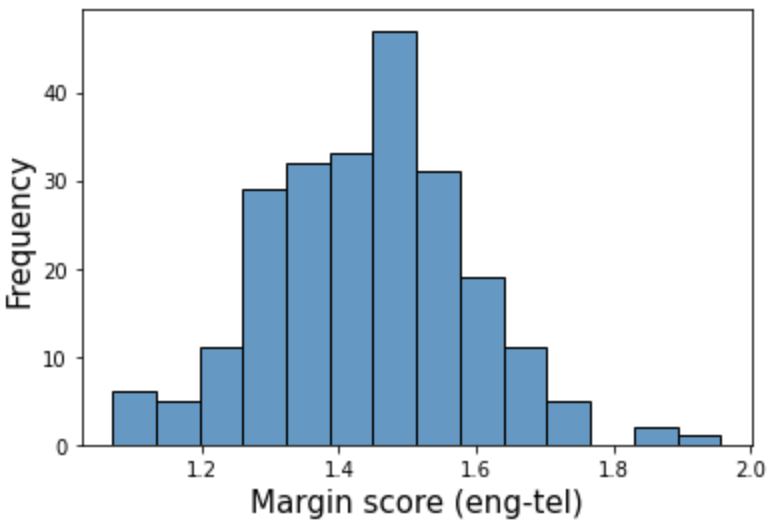}}
\caption{Distributions of margin scores across various datasets, achieved using \textit{intersection} retrieval with no threshold. The left three graphs are mined from comparable corpora, while the right two are mined from gold-standard bitexts and contain 99\%+ ground-truth pairs.}
\label{fig:margin-plots}
\end{figure*}

One benefit of our proposed approach is that it is threshold-agnostic, unlike previous approaches \citep{artetxe-schwenk-2019-margin, wikimatrix, ccmatrix}. Furthermore, our results on semi-supervised (really, supervised+unsupervised+semi-supervised) MT show that adding data mined using the pairwise intersection method (Method \textbf{2} in Table \ref{tab:bleu-scores}) improves over the WMT'19 baseline, while adding data mined using a threshold of 1.2 actually \textit{hurts} performance considerably.
These results are in line with the somewhat arbitrary nature of margin score thresholding observed elsewhere. Figure \ref{fig:margin-plots} shows distributions of margin scores on sentence pairs mined on our English-Kazakh comparable corpus (using document-level mining), on the BUCC English-French training data (globally mined) and on two Tatoeba test sets, namely English-Maltese and English-Telugu (also globally mined). \\
First, we note that the margin distributions on the latter two datasets—for which we've plotted 99\%+ ground-truth pairs—appear approximately normally distributed over a significantly large range (around size 0.7-1 for both), rendering it impossible to choose a single threshold that catches all pairs. This is in line with the much more extensive results displayed in Table \ref{table_tatoeba_all}, in which only a few of the 64 language pairs aren't harmed by even a low threshold of 1.06. On the BUCC data, the margin scores appear almost perfectly normally distributed, seeming to belie our critique. However, a close analysis of this distribution reveals a small local maximum around 1.3, most likely representing the gold-standard pairs that were injected into the BUCC corpus \citep{zweigenbaum-etal-2017-overview, zweigenbaum:hal-01898360}. This may explain the success of this threshold in others' studies using this dataset \footnote{\href{https://www.sbert.net/examples/applications/parallel-sentence-mining/README.html}{https://www.sbert.net/examples/applications/parallel-sentence-mining/README.html}}. \\
Another issue is that the optimal margin threshold appears dependent on the size of the search space, posing a particular issue for document-level mining in which this size differs from document to document. The choice of margin threshold is discussed in both \citet{ccmatrix} and \citet{wikimatrix}, but neither address the topic with much rigor. \citet{wikimatrix} examines a very narrow range of margin thresholds for only two language pairs (four directions) on bitexts mined from Wikipedia, but yield no truly conclusive results, nor any disciplined method for selecting the optimal threshold. The same may be said of \citet{ccmatrix}, in which the optimal threshold is justified by BLEU evaluation on a single language pair. The margin score \citet{ccmatrix} uses to mine globally over millions or billions of sentences performs sub-optimally on our corpus using document level mining, with higher thresholds yielding only marginal improvements.
To the contrary, we speculate that the various signals in our voting-based approach provide the same sort of denoising effect as other voting-based approaches (e.g. voting models), and while Table \ref{tab:bleu-scores} shows that the generated bitexts still benefit from thresholding (see results under "Previous training procedure," which simply involved training for less time on less GPUs), voting alone acts as a sufficient heuristic to produce reasonably good precision and recall. As can be seen by Figure \ref{fig:margin-mv}, the proposed approach does \textit{not} perform a sort of implicit thresholding.


\section{Discussion}

\subsection{Cross-lingual Alignment in Multilingual Sentence Embedding Models}

One upshot of the results of our approach is that evidently, there is something to be gained from translating texts before performing similarity search, that there is something salient in this signal that is not in the signal generated by the original text's embedding. What this points to is some deficiency in the cross-lingual alignment in LaBSE (and likely in other cross-lingual sentence embedding models as well). \\
Cross-lingual alignment has been investigated in the context of monolingual embeddings rather rigorously. \citet{vulic-etal-2020-good} investigates the causes of cross-lingual misalignment between monolingual embedding spaces, and pinpoints language model training data size and training regimes as the main culprits, to the exclusion of typological factors such as morphology and word order. Furthermore, \citet{pires-etal-2019-multilingual} and \citet{wu-dredze-2019-beto} investigate the cross-lingual alignment ability of mBERT \citet{devlin-etal-2019-bert} by examining the zero-shot case (i.e. fine-tuning on one language and predicting on others) for a variety of tasks. However, while \citet{pires-etal-2019-multilingual, wu-dredze-2019-beto} each touch on linguistic factors, neither thoroughly investigate their effect on cross-lingual \textit{alignment} (as opposed to \textit{transfer}, which we argue may be correlated but not perfectly so), or make a rigorous effort to control for the other factors at play in cross-lingual LMs, such as monolingual/bilingual training data size and size of same-family training data for a given language. While such an investigation lies beyond the scope of this paper, we believe our results make practical use of a deficient cross-lingual alignment, and echo \citet{artetxe-etal-2020-call}'s call for more thorough probing, linguistic and otherwise, of cross-lingual models.

\subsection{Energy and Resource Considerations}

While catalyzed by such tools as FAISS \citep{faiss}, bitext mining is inherently an incredibly expensive task, especially when performed globally. Though extensive computational optimization has allowed the search space to grow to billions of sentences \citep{fan2020englishcentric}, these global-level procedures still require hundreds of GPUs, leaving a sizable environmental footprint \citep{green-ai, strubell-etal-2019-energy} and limiting the number of researchers and institutions to whom this method is available.
Our approach, while requiring the upfront cost of translating entire corpora of sentence pairs, operates at the document level (which, as \citet{ccmatrix} note, is available for Wikipedia but not for corpora like Common Crawl) and provides a heuristic measure that may eliminate the need for laboriously tuning a somewhat arbitrary margin threshold.

\subsection{Linguistic Diversity}

\citet{call-for-rigor} outlines many of the key issues in unsupervised cross-lingual learning and evaluation, among which are the verisimilitude of the training conditions and the lack of a cross-lingual benchmark for many tasks. On the one hand, our method relies on supervised data and thus isn't applicable to the most low-resource language pairs, helping instead a niche of mid-low-resource languages (see Table \ref{tatoeba-average}). Also, we report results on the Tatoeba test set from \citet{artetxe-schwenk-2019-laser}, which contains only English-aligned sentence pairs. However, as \citet{call-for-rigor} note, the fully supervised setting isn't as rare as it's often made out to be, and our relatively lightweight, heuristic-based approach suits a practical research or development setting. Nonetheless, we would like to perform extensive linguistic probing on the bitext mining task using non-English-aligned corpora, and suggest \citet{tiedemann-2020-tatoeba} as a possible resource for this inquiry, as it extends the Tatoeba test set from \citet{artetxe-schwenk-2019-laser} to nearly 3000 language pairs. As \citet{ccmatrix} note, the factors affecting bitext quality and quantity aren't fully understood. Such massively multilingual, non-English-centric benchmarks will enable richer and more inclusive cross-lingual research \citep{joshi-etal-2020-state}, building on top of current benchmarks such as XTREME, XGLUE, and XNLI \citep{xtreme, xglue, xnli}, and supplementing probing efforts such as \citet{pires-etal-2019-multilingual} and \citet{wu-dredze-2019-beto}.


\section{Conclusions}

In this paper, we propose a novel method of mining sentence pairs from both comparable and parallel corpora, and demonstrate success on both the Tatoeba gold-standard bitext mining task and on mining pseudo-parallel sentences for NMT. We uncover the problematic nature of setting a similarity score threshold for this task, particularly in the context of margin scoring with document-level mining, showing that thresholding is not a one-size-fits-all approach. On the Tatoeba dataset, we set what we believe to be new benchmarks for 64 languages and reveal an intriguing cross-lingual division across languages by their resource availability with respect to which mining approach performs best, with the voting-based approach involving bidirectional translation providing superior results on languages for which a supervised NMT system was available. We contribute novel insights regarding the cross-lingual alignment of multilingual language models, exploit its deficiencies, and propose further probing efforts to examine the linguistic and technical factors affecting this alignment. In future work, we also hope to investigate how cross-lingual alignment may be improved in cross-lingual LMs, and how our mining methods transfer to the large-scale, global mining scenario.

\section{Acknowledgements}

We would like to thank Mohammad Sadegh Rasooli for offering ideas and insights, as well as Nils Reimers, Iryna Gurevych, and the Facebook AI team for contributing their open-source models and data for research. We would also like to acknowledge Boston University for providing funding and computational resources.

\bibliography{anthology,emnlp2020}

\begin{thebibliography}{85}
\expandafter\ifx\csname natexlab\endcsname\relax\def\natexlab#1{#1}\fi

\bibitem[{uns(2021)}]{unsupMTuseful2021}
 2021.
\newblock {Unsupervised Machine Translation is Useful: It Just Needs Good
  Company}.

\bibitem[{{Ammar} et~al.(2016){Ammar}, {Mulcaire}, {Tsvetkov}, {Lample},
  {Dyer}, and {Smith}}]{ammar-2016}
Waleed {Ammar}, George {Mulcaire}, Yulia {Tsvetkov}, Guillaume {Lample}, Chris
  {Dyer}, and Noah~A. {Smith}. 2016.
\newblock \href {http://arxiv.org/abs/1602.01925} {{Massively Multilingual Word
  Embeddings}}.
\newblock \emph{arXiv e-prints, arXiv:1602.01925}.

\bibitem[{Arora et~al.(2017)Arora, Liang, and Ma}]{simple-baseline}
Sanjeev Arora, Yingyu Liang, and Tengyu Ma. 2017.
\newblock \href {https://openreview.net/pdf?id=SyK00v5xx} {A simple but
  tough-to-beat baseline for sentence embeddings}.
\newblock \emph{International Conference on Learning Representations 2017}.

\bibitem[{Artetxe et~al.(2018{\natexlab{a}})Artetxe, Labaka, and
  Agirre}]{artetxe-etal-2018-robust2}
Mikel Artetxe, Gorka Labaka, and Eneko Agirre. 2018{\natexlab{a}}.
\newblock \href {https://doi.org/10.18653/v1/P18-1073} {{A Robust Self-Learning
  Method for Fully Unsupervised Cross-Lingual Mappings of Word Embeddings}}.
\newblock In \emph{Proceedings of the 56th Annual Meeting of the Association
  for Computational Linguistics (Volume 1: Long Papers)}, pages 789--798,
  Melbourne, Australia. Association for Computational Linguistics.

\bibitem[{Artetxe et~al.(2018{\natexlab{b}})Artetxe, Labaka, and
  Agirre}]{artetxe2018emnlp}
Mikel Artetxe, Gorka Labaka, and Eneko Agirre. 2018{\natexlab{b}}.
\newblock \href {https://www.aclweb.org/anthology/D18-1399} {{Unsupervised
  Statistical Machine Translation}}.
\newblock In \emph{Proceedings of the 2018 Conference on Empirical Methods in
  Natural Language Processing}, pages 3632--3642, Brussels, Belgium.
  Association for Computational Linguistics.

\bibitem[{Artetxe et~al.(2019{\natexlab{a}})Artetxe, Labaka, and
  Agirre}]{artetxe2019acl-umt}
Mikel Artetxe, Gorka Labaka, and Eneko Agirre. 2019{\natexlab{a}}.
\newblock \href {https://www.aclweb.org/anthology/P19-1019} {{An Effective
  Approach to Unsupervised Machine Translation}}.
\newblock In \emph{Proceedings of the 57th Annual Meeting of the Association
  for Computational Linguistics}, pages 194--203, Florence, Italy. Association
  for Computational Linguistics.

\bibitem[{Artetxe et~al.(2019{\natexlab{b}})Artetxe, Labaka, and
  Agirre}]{artetxe2019acl-bli}
Mikel Artetxe, Gorka Labaka, and Eneko Agirre. 2019{\natexlab{b}}.
\newblock \href {https://www.aclweb.org/anthology/P19-1494} {{Bilingual Lexicon
  Induction through Unsupervised Machine Translation}}.
\newblock In \emph{Proceedings of the 57th Annual Meeting of the Association
  for Computational Linguistics}, pages 5002--5007, Florence, Italy.
  Association for Computational Linguistics.

\bibitem[{Artetxe et~al.(2018{\natexlab{c}})Artetxe, Labaka, Agirre, and
  Cho}]{artetxe2018unsupervised}
Mikel Artetxe, Gorka Labaka, Eneko Agirre, and Kyunghyun Cho.
  2018{\natexlab{c}}.
\newblock \href {https://openreview.net/forum?id=Sy2ogebAW} {{Unsupervised
  Neural Machine Translation}}.
\newblock In \emph{International Conference on Learning Representations 2018}.

\bibitem[{{Artetxe} et~al.(2020){Artetxe}, {Ruder}, {Yogatama}, {Labaka}, and
  {Agirre}}]{call-for-rigor}
Mikel {Artetxe}, Sebastian {Ruder}, Dani {Yogatama}, Gorka {Labaka}, and Eneko
  {Agirre}. 2020.
\newblock \href {http://arxiv.org/abs/2004.14958} {{A Call for More Rigor in
  Unsupervised Cross-lingual Learning}}.
\newblock \emph{arXiv e-prints}, page arXiv:2004.14958.

\bibitem[{Artetxe et~al.(2020)Artetxe, Ruder, Yogatama, Labaka, and
  Agirre}]{artetxe-etal-2020-call}
Mikel Artetxe, Sebastian Ruder, Dani Yogatama, Gorka Labaka, and Eneko Agirre.
  2020.
\newblock \href {https://doi.org/10.18653/v1/2020.acl-main.658} {A call for
  more rigor in unsupervised cross-lingual learning}.
\newblock In \emph{Proceedings of the 58th Annual Meeting of the Association
  for Computational Linguistics}, pages 7375--7388, Online. Association for
  Computational Linguistics.

\bibitem[{Artetxe and
  Schwenk(2019{\natexlab{a}})}]{artetxe-schwenk-2019-margin}
Mikel Artetxe and Holger Schwenk. 2019{\natexlab{a}}.
\newblock \href {https://doi.org/10.18653/v1/P19-1309} {Margin-based parallel
  corpus mining with multilingual sentence embeddings}.
\newblock In \emph{Proceedings of the 57th Annual Meeting of the Association
  for Computational Linguistics}, pages 3197--3203, Florence, Italy.
  Association for Computational Linguistics.

\bibitem[{Artetxe and Schwenk(2019{\natexlab{b}})}]{artetxe-schwenk-2019-laser}
Mikel Artetxe and Holger Schwenk. 2019{\natexlab{b}}.
\newblock \href {https://doi.org/10.1162/tacl\_a\_00288} {{Massively
  Multilingual Sentence Embeddings for Zero-Shot Cross-Lingual Transfer and
  Beyond}}.
\newblock \emph{Transactions of the Association for Computational Linguistics},
  7:597--610.

\bibitem[{Azpeitia et~al.(2018)Azpeitia, Etchegoyhen, and
  Garcia}]{azpeitia-stacc}
Adoni Azpeitia, Thierry Etchegoyhen, and Eva~Martinez Garcia. 2018.
\newblock \href {http://lrec-conf.org/workshops/lrec2018/W8/pdf/6_W8.pdf}
  {{Extracting Parallel Sentences from Comparable Corpora with {STACC}
  Variants}}.
\newblock In \emph{Proceedings of the Eleventh International Conference on
  Language Resources and Evaluation (LREC 2018)}, Paris, France. European
  Language Resources Association (ELRA).

\bibitem[{Azpeitia et~al.(2017)Azpeitia, Etchegoyhen, and
  Mart{\'\i}nez~Garcia}]{azpeitia-etal-2017-weighted}
Andoni Azpeitia, Thierry Etchegoyhen, and Eva Mart{\'\i}nez~Garcia. 2017.
\newblock \href {https://doi.org/10.18653/v1/W17-2508} {Weighted set-theoretic
  alignment of comparable sentences}.
\newblock In \emph{Proceedings of the 10th Workshop on Building and Using
  Comparable Corpora}, pages 41--45, Vancouver, Canada. Association for
  Computational Linguistics.

\bibitem[{Barrault et~al.(2019)Barrault, Bojar, Costa-juss{\`a}, Federmann,
  Fishel, Graham, Haddow, Huck, Koehn, Malmasi, Monz, M{\"u}ller, Pal, Post,
  and Zampieri}]{wmt2019}
Lo{\"\i}c Barrault, Ond{\v{r}}ej Bojar, Marta~R. Costa-juss{\`a}, Christian
  Federmann, Mark Fishel, Yvette Graham, Barry Haddow, Matthias Huck, Philipp
  Koehn, Shervin Malmasi, Christof Monz, Mathias M{\"u}ller, Santanu Pal, Matt
  Post, and Marcos Zampieri. 2019.
\newblock \href {https://doi.org/10.18653/v1/W19-5301} {Findings of the 2019
  conference on machine translation ({WMT}19)}.
\newblock In \emph{Proceedings of the Fourth Conference on Machine Translation
  (Volume 2: Shared Task Papers, Day 1)}, pages 1--61, Florence, Italy.
  Association for Computational Linguistics.

\bibitem[{Bouamor and Sajjad(2018)}]{BOUAMOR18.8}
Houda Bouamor and Hassan Sajjad. 2018.
\newblock \href {http://lrec-conf.org/workshops/lrec2018/W8/pdf/8_W8.pdf}
  {{H2@BUCC18: Parallel Sentence Extraction from Comparable Corpora Using
  Multilingual Sentence Embeddings}}.
\newblock In \emph{Proceedings of the Eleventh International Conference on
  Language Resources and Evaluation (LREC 2018)}, Paris, France. European
  Language Resources Association (ELRA).

\bibitem[{Butz and Wilson(2002)}]{butz-wilson-xcs}
M.V. Butz and S.W. Wilson. 2002.
\newblock \href {https://doi.org/10.1007/s005000100111} {{An Algorithmic
  Description of XCS}}.
\newblock \emph{Soft Computing}, 6:144--153.

\bibitem[{Cao et~al.(2020)Cao, Kitaev, and Klein}]{Cao2020Multilingual}
Steven Cao, Nikita Kitaev, and Dan Klein. 2020.
\newblock \href {https://openreview.net/forum?id=r1xCMyBtPS} {{Multilingual
  Alignment of Contextual Word Representations}}.
\newblock In \emph{International Conference on Learning Representations}.

\bibitem[{Choe et~al.(2020)Choe, Park, and Kim}]{choe-etal-2020-word2word}
Yo~Joong Choe, Kyubyong Park, and Dongwoo Kim. 2020.
\newblock \href {https://www.aclweb.org/anthology/2020.lrec-1.371} {{word2word:
  A Collection of Bilingual Lexicons for 3,564 Language Pairs}}.
\newblock In \emph{Proceedings of the 12th Language Resources and Evaluation
  Conference}, pages 3036--3045, Marseille, France. European Language Resources
  Association.

\bibitem[{Conneau et~al.(2020)Conneau, Khandelwal, Goyal, Chaudhary, Wenzek,
  Guzm{\'a}n, Grave, Ott, Zettlemoyer, and
  Stoyanov}]{conneau-etal-2020-unsupervised}
Alexis Conneau, Kartikay Khandelwal, Naman Goyal, Vishrav Chaudhary, Guillaume
  Wenzek, Francisco Guzm{\'a}n, Edouard Grave, Myle Ott, Luke Zettlemoyer, and
  Veselin Stoyanov. 2020.
\newblock \href {https://doi.org/10.18653/v1/2020.acl-main.747} {{Unsupervised
  Cross-lingual Representation Learning at Scale}}.
\newblock In \emph{Proceedings of the 58th Annual Meeting of the Association
  for Computational Linguistics}, pages 8440--8451, Online. Association for
  Computational Linguistics.

\bibitem[{{Conneau} et~al.(2017){Conneau}, {Kiela}, {Schwenk}, {Barrault}, and
  {Bordes}}]{infersent}
Alexis {Conneau}, Douwe {Kiela}, Holger {Schwenk}, Loic {Barrault}, and Antoine
  {Bordes}. 2017.
\newblock \href {http://arxiv.org/abs/1705.02364} {{Supervised Learning of
  Universal Sentence Representations from Natural Language Inference Data}}.
\newblock \emph{arXiv e-prints}, page arXiv:1705.02364.

\bibitem[{Conneau and Lample(2019{\natexlab{a}})}]{NEURIPS2019_c04c19c2}
Alexis Conneau and Guillaume Lample. 2019{\natexlab{a}}.
\newblock \href
  {https://proceedings.neurips.cc/paper/2019/file/c04c19c2c2474dbf5f7ac4372c5b9af1-Paper.pdf}
  {{Cross-lingual Language Model Pretraining}}.
\newblock In \emph{Advances in Neural Information Processing Systems},
  volume~32. Curran Associates, Inc.

\bibitem[{Conneau and Lample(2019{\natexlab{b}})}]{conneau2019cross}
Alexis Conneau and Guillaume Lample. 2019{\natexlab{b}}.
\newblock Cross-lingual language model pretraining.
\newblock In \emph{Advances in Neural Information Processing Systems}, pages
  7059--7069.

\bibitem[{{Conneau} et~al.(2018){Conneau}, {Lample}, {Rinott}, {Williams},
  {Bowman}, {Schwenk}, and {Stoyanov}}]{xnli}
Alexis {Conneau}, Guillaume {Lample}, Ruty {Rinott}, Adina {Williams},
  Samuel~R. {Bowman}, Holger {Schwenk}, and Veselin {Stoyanov}. 2018.
\newblock \href {http://arxiv.org/abs/1809.05053} {{XNLI: Evaluating
  Cross-lingual Sentence Representations}}.
\newblock \emph{arXiv e-prints}, page arXiv:1809.05053.

\bibitem[{Devlin et~al.(2019)Devlin, Chang, Lee, and
  Toutanova}]{devlin-etal-2019-bert}
Jacob Devlin, Ming-Wei Chang, Kenton Lee, and Kristina Toutanova. 2019.
\newblock \href {https://doi.org/10.18653/v1/N19-1423} {{BERT}: Pre-training of
  deep bidirectional transformers for language understanding}.
\newblock In \emph{Proceedings of the 2019 Conference of the North {A}merican
  Chapter of the Association for Computational Linguistics: Human Language
  Technologies, Volume 1 (Long and Short Papers)}, pages 4171--4186,
  Minneapolis, Minnesota. Association for Computational Linguistics.

\bibitem[{{Dinu} et~al.(2014){Dinu}, {Lazaridou}, and {Baroni}}]{hubness}
Georgiana {Dinu}, Angeliki {Lazaridou}, and Marco {Baroni}. 2014.
\newblock \href {http://arxiv.org/abs/1412.6568} {{Improving zero-shot learning
  by mitigating the hubness problem}}.
\newblock \emph{arXiv e-prints}, page arXiv:1412.6568.

\bibitem[{Etchegoyhen and Azpeitia(2016)}]{etchegoyhen-azpeitia-2016-set}
Thierry Etchegoyhen and Andoni Azpeitia. 2016.
\newblock \href {https://doi.org/10.18653/v1/P16-1189} {Set-theoretic alignment
  for comparable corpora}.
\newblock In \emph{Proceedings of the 54th Annual Meeting of the Association
  for Computational Linguistics (Volume 1: Long Papers)}, pages 2009--2018,
  Berlin, Germany. Association for Computational Linguistics.

\bibitem[{{Fan} et~al.(2020){Fan}, {Bhosale}, {Schwenk}, {Ma}, {El-Kishky},
  {Goyal}, {Baines}, {Celebi}, {Wenzek}, {Chaudhary}, {Goyal}, {Birch},
  {Liptchinsky}, {Edunov}, {Grave}, {Auli}, and
  {Joulin}}]{fan2020englishcentric}
Angela {Fan}, Shruti {Bhosale}, Holger {Schwenk}, Zhiyi {Ma}, Ahmed
  {El-Kishky}, Siddharth {Goyal}, Mandeep {Baines}, Onur {Celebi}, Guillaume
  {Wenzek}, Vishrav {Chaudhary}, Naman {Goyal}, Tom {Birch}, Vitaliy
  {Liptchinsky}, Sergey {Edunov}, Edouard {Grave}, Michael {Auli}, and Armand
  {Joulin}. 2020.
\newblock \href {http://arxiv.org/abs/2010.11125} {{Beyond English-Centric
  Multilingual Machine Translation}}.
\newblock \emph{arXiv e-prints}, page arXiv:2010.11125.

\bibitem[{{Feng} et~al.(2020){Feng}, {Yang}, {Cer}, {Arivazhagan}, and
  {Wang}}]{feng2020labse}
Fangxiaoyu {Feng}, Yinfei {Yang}, Daniel {Cer}, Naveen {Arivazhagan}, and Wei
  {Wang}. 2020.
\newblock \href {http://arxiv.org/abs/2007.01852} {{Language-agnostic BERT
  Sentence Embedding}}.
\newblock \emph{arXiv e-prints}, page arXiv:2007.01852.

\bibitem[{Finkel et~al.(2005)Finkel, Grenager, and
  Manning}]{finkel-etal-2005-incorporating}
Jenny~Rose Finkel, Trond Grenager, and Christopher Manning. 2005.
\newblock \href {https://doi.org/10.3115/1219840.1219885} {Incorporating
  non-local information into information extraction systems by {G}ibbs
  sampling}.
\newblock In \emph{Proceedings of the 43rd Annual Meeting of the Association
  for Computational Linguistics ({ACL}{'}05)}, pages 363--370, Ann Arbor,
  Michigan. Association for Computational Linguistics.

\bibitem[{Fung and Cheung(2004)}]{fung-cheung-2004-multi}
Pascale Fung and Percy Cheung. 2004.
\newblock \href {https://www.aclweb.org/anthology/C04-1151} {Multi-level
  bootstrapping for extracting parallel sentences from a quasi-comparable
  corpus}.
\newblock In \emph{{COLING} 2004: Proceedings of the 20th International
  Conference on Computational Linguistics}, pages 1051--1057, Geneva,
  Switzerland. COLING.

\bibitem[{Hangya et~al.(2018)Hangya, Braune, Kalasouskaya, and
  Fraser}]{Hangya2018Unsupervised}
Viktor Hangya, Fabienne Braune, Yuliya Kalasouskaya, and Alexander Fraser.
  2018.
\newblock \href
  {https://www.cis.uni-muenchen.de/~fraser/pubs/hangya_iwslt2018.pdf}
  {{Unsupervised Parallel Sentence Extraction from Comparable Corpora}}.
\newblock In \emph{Proceedings of the 15th International Workshop on Spoken
  Language Translation}, pages 7--13, Bruges, Belgium.

\bibitem[{Hangya and Fraser(2019)}]{hangya-fraser-2019-unsupervised}
Viktor Hangya and Alexander Fraser. 2019.
\newblock \href {https://doi.org/10.18653/v1/P19-1118} {Unsupervised parallel
  sentence extraction with parallel segment detection helps machine
  translation}.
\newblock In \emph{Proceedings of the 57th Annual Meeting of the Association
  for Computational Linguistics}, pages 1224--1234, Florence, Italy.
  Association for Computational Linguistics.

\bibitem[{Hoang et~al.(2018)Hoang, Koehn, Haffari, and
  Cohn}]{hoang-etal-2018-iterative}
Vu~Cong~Duy Hoang, Philipp Koehn, Gholamreza Haffari, and Trevor Cohn. 2018.
\newblock \href {https://doi.org/10.18653/v1/W18-2703} {Iterative
  back-translation for neural machine translation}.
\newblock In \emph{Proceedings of the 2nd Workshop on Neural Machine
  Translation and Generation}, pages 18--24, Melbourne, Australia. Association
  for Computational Linguistics.

\bibitem[{{Hu} et~al.(2020){Hu}, {Ruder}, {Siddhant}, {Neubig}, {Firat}, and
  {Johnson}}]{xtreme}
Junjie {Hu}, Sebastian {Ruder}, Aditya {Siddhant}, Graham {Neubig}, Orhan
  {Firat}, and Melvin {Johnson}. 2020.
\newblock \href {http://arxiv.org/abs/2003.11080} {{XTREME: A Massively
  Multilingual Multi-task Benchmark for Evaluating Cross-lingual
  Generalization}}.
\newblock \emph{arXiv e-prints}, page arXiv:2003.11080.

\bibitem[{{Johnson} et~al.(2017){Johnson}, {Douze}, and {J{\'e}gou}}]{faiss}
Jeff {Johnson}, Matthijs {Douze}, and Herv{\'e} {J{\'e}gou}. 2017.
\newblock \href {http://arxiv.org/abs/1702.08734} {{Billion-scale similarity
  search with GPUs}}.
\newblock \emph{arXiv e-prints}, page arXiv:1702.08734.

\bibitem[{Joshi et~al.(2020)Joshi, Santy, Budhiraja, Bali, and
  Choudhury}]{joshi-etal-2020-state}
Pratik Joshi, Sebastin Santy, Amar Budhiraja, Kalika Bali, and Monojit
  Choudhury. 2020.
\newblock \href {https://doi.org/10.18653/v1/2020.acl-main.560} {{The State and
  Fate of Linguistic Diversity and Inclusion in the {NLP} World}}.
\newblock In \emph{Proceedings of the 58th Annual Meeting of the Association
  for Computational Linguistics}, pages 6282--6293, Online. Association for
  Computational Linguistics.

\bibitem[{Keung et~al.(2020)Keung, Salazar, Lu, and
  Smith}]{keung2020unsupervised}
Phillip Keung, Julian Salazar, Yichao Lu, and Noah~A. Smith. 2020.
\newblock \href {http://arxiv.org/abs/2010.07761} {Unsupervised {B}itext
  {M}ining and {T}ranslation via {S}elf-trained {C}ontextual {E}mbeddings}.

\bibitem[{Kim et~al.(2020)Kim, Gra{\c{c}}a, and
  Ney}]{kim-etal-2020-unsupervised}
Yunsu Kim, Miguel Gra{\c{c}}a, and Hermann Ney. 2020.
\newblock \href {https://www.aclweb.org/anthology/2020.eamt-1.5} {{When and Why
  is Unsupervised Neural Machine Translation Useless?}}
\newblock In \emph{Proceedings of the 22nd Annual Conference of the European
  Association for Machine Translation}, pages 35--44, Lisboa, Portugal.
  European Association for Machine Translation.

\bibitem[{{Kiros} et~al.(2015){Kiros}, {Zhu}, {Salakhutdinov}, {Zemel},
  {Torralba}, {Urtasun}, and {Fidler}}]{skip-thought}
Ryan {Kiros}, Yukun {Zhu}, Ruslan {Salakhutdinov}, Richard~S. {Zemel}, Antonio
  {Torralba}, Raquel {Urtasun}, and Sanja {Fidler}. 2015.
\newblock \href {http://arxiv.org/abs/1506.06726} {{Skip-Thought Vectors}}.
\newblock \emph{arXiv e-prints}, page arXiv:1506.06726.

\bibitem[{Kitaev et~al.(2019)Kitaev, Cao, and
  Klein}]{kitaev-etal-2019-multilingual}
Nikita Kitaev, Steven Cao, and Dan Klein. 2019.
\newblock \href {https://doi.org/10.18653/v1/P19-1340} {Multilingual
  constituency parsing with self-attention and pre-training}.
\newblock In \emph{Proceedings of the 57th Annual Meeting of the Association
  for Computational Linguistics}, pages 3499--3505, Florence, Italy.
  Association for Computational Linguistics.

\bibitem[{Klementiev et~al.(2012)Klementiev, Titov, and
  Bhattarai}]{klementiev-etal-2012-inducing}
Alexandre Klementiev, Ivan Titov, and Binod Bhattarai. 2012.
\newblock \href {https://www.aclweb.org/anthology/C12-1089} {Inducing
  crosslingual distributed representations of words}.
\newblock In \emph{Proceedings of {COLING} 2012}, pages 1459--1474, Mumbai,
  India. The COLING 2012 Organizing Committee.

\bibitem[{Kvapil{\'\i}kov{\'a} et~al.(2020)Kvapil{\'\i}kov{\'a}, Artetxe,
  Labaka, Agirre, and Bojar}]{kvapilikova-etal-2020-unsupervised}
Ivana Kvapil{\'\i}kov{\'a}, Mikel Artetxe, Gorka Labaka, Eneko Agirre, and
  Ond{\v{r}}ej Bojar. 2020.
\newblock \href {https://doi.org/10.18653/v1/2020.acl-srw.34} {Unsupervised
  {M}ultilingual {S}entence {E}mbeddings for {P}arallel {C}orpus {M}ining}.
\newblock In \emph{Proceedings of the 58th Annual Meeting of the Association
  for Computational Linguistics: Student Research Workshop}, pages 255--262,
  Online. Association for Computational Linguistics.

\bibitem[{{Lample} et~al.(2017){Lample}, {Conneau}, {Denoyer}, and
  {Ranzato}}]{lample-2017-unmt}
Guillaume {Lample}, Alexis {Conneau}, Ludovic {Denoyer}, and Marc'Aurelio
  {Ranzato}. 2017.
\newblock \href {http://arxiv.org/abs/1711.00043} {{Unsupervised Machine
  Translation Using Monolingual Corpora Only}}.
\newblock \emph{arXiv e-prints}, page arXiv:1711.00043.

\bibitem[{Lample et~al.(2018{\natexlab{a}})Lample, Conneau, Denoyer, and
  Ranzato}]{lample2018unsupervised}
Guillaume Lample, Alexis Conneau, Ludovic Denoyer, and Marc'Aurelio Ranzato.
  2018{\natexlab{a}}.
\newblock \href {https://openreview.net/forum?id=rkYTTf-AZ} {Unsupervised
  machine translation using monolingual corpora only}.
\newblock In \emph{International Conference on Learning Representations}.

\bibitem[{Lample et~al.(2018{\natexlab{b}})Lample, Conneau, Ranzato, Denoyer,
  and Jégou}]{lample2018word}
Guillaume Lample, Alexis Conneau, Marc'Aurelio Ranzato, Ludovic Denoyer, and
  Hervé Jégou. 2018{\natexlab{b}}.
\newblock \href {https://openreview.net/forum?id=H196sainb} {{Word translation
  without parallel data}}.
\newblock In \emph{International Conference on Learning Representations}.

\bibitem[{Lample et~al.(2018{\natexlab{c}})Lample, Ott, Conneau, Denoyer, and
  Ranzato}]{lample-etal-2018-phrase}
Guillaume Lample, Myle Ott, Alexis Conneau, Ludovic Denoyer, and Marc{'}Aurelio
  Ranzato. 2018{\natexlab{c}}.
\newblock \href {https://doi.org/10.18653/v1/D18-1549} {Phrase-based {\&}
  neural unsupervised machine translation}.
\newblock In \emph{Proceedings of the 2018 Conference on Empirical Methods in
  Natural Language Processing}, pages 5039--5049, Brussels, Belgium.
  Association for Computational Linguistics.

\bibitem[{{Liang} et~al.(2020){Liang}, {Duan}, {Gong}, {Wu}, {Guo}, {Qi},
  {Gong}, {Shou}, {Jiang}, {Cao}, {Fan}, {Zhang}, {Agrawal}, {Cui}, {Wei},
  {Bharti}, {Qiao}, {Chen}, {Wu}, {Liu}, {Yang}, {Campos}, {Majumder}, and
  {Zhou}}]{xglue}
Yaobo {Liang}, Nan {Duan}, Yeyun {Gong}, Ning {Wu}, Fenfei {Guo}, Weizhen {Qi},
  Ming {Gong}, Linjun {Shou}, Daxin {Jiang}, Guihong {Cao}, Xiaodong {Fan},
  Ruofei {Zhang}, Rahul {Agrawal}, Edward {Cui}, Sining {Wei}, Taroon {Bharti},
  Ying {Qiao}, Jiun-Hung {Chen}, Winnie {Wu}, Shuguang {Liu}, Fan {Yang},
  Daniel {Campos}, Rangan {Majumder}, and Ming {Zhou}. 2020.
\newblock \href {http://arxiv.org/abs/2004.01401} {{XGLUE: A New Benchmark
  Dataset for Cross-lingual Pre-training, Understanding and Generation}}.
\newblock \emph{arXiv e-prints}, page arXiv:2004.01401.

\bibitem[{Liu et~al.(2020)Liu, Ott, Goyal, Du, Joshi, Chen, Levy, Lewis,
  Zettlemoyer, and Stoyanov}]{liu2020roberta}
Yinhan Liu, Myle Ott, Naman Goyal, Jingfei Du, Mandar Joshi, Danqi Chen, Omer
  Levy, Mike Lewis, Luke Zettlemoyer, and Veselin Stoyanov. 2020.
\newblock \href {https://openreview.net/forum?id=SyxS0T4tvS} {{Ro{\{}BERT{\}}a:
  A Robustly Optimized {\{}BERT{\}} Pretraining Approach}}.

\bibitem[{Lukins(2002)}]{lukins-2002}
Timothy~Campbell Lukins. 2002.
\newblock \emph{{Dynamically Developing Novel and Useful Behaviours: A First
  Step in Animat Creativity}}.
\newblock Ph.D. thesis.

\bibitem[{Marchisio et~al.(2020)Marchisio, Duh, and
  Koehn}]{marchisio-etal-2020-unsupervised}
Kelly Marchisio, Kevin Duh, and Philipp Koehn. 2020.
\newblock \href {https://www.aclweb.org/anthology/2020.wmt-1.68} {{When Does
  Unsupervised Machine Translation Work?}}
\newblock In \emph{Proceedings of the Fifth Conference on Machine Translation},
  pages 571--583, Online. Association for Computational Linguistics.

\bibitem[{Mikolov et~al.(2013)Mikolov, Sutskever, Chen, Corrado, and
  Dean}]{mikolov-etal-dist}
Tomas Mikolov, Ilya Sutskever, Kai Chen, Greg Corrado, and Jeffrey Dean. 2013.
\newblock \href
  {https://papers.nips.cc/paper/2013/file/9aa42b31882ec039965f3c4923ce901b-Paper.pdf}
  {{Distributed Representations of Words and Phrases and Their
  Compositionality}}.
\newblock In \emph{Proceedings of the 26th International Conference on Neural
  Information Processing Systems - Volume 2}, NIPS'13, page 3111–3119, Red
  Hook, NY, USA. Curran Associates Inc.

\bibitem[{Munteanu et~al.(2004)Munteanu, Fraser, and
  Marcu}]{munteanu-etal-2004-improved}
Dragos~Stefan Munteanu, Alexander Fraser, and Daniel Marcu. 2004.
\newblock \href {https://www.aclweb.org/anthology/N04-1034} {Improved machine
  translation performance via parallel sentence extraction from comparable
  corpora}.
\newblock In \emph{Proceedings of the Human Language Technology Conference of
  the North {A}merican Chapter of the Association for Computational
  Linguistics: {HLT}-{NAACL} 2004}, pages 265--272, Boston, Massachusetts, USA.
  Association for Computational Linguistics.

\bibitem[{Munteanu and Marcu(2006)}]{munteanu-marcu-2006-extracting}
Dragos~Stefan Munteanu and Daniel Marcu. 2006.
\newblock \href {https://doi.org/10.3115/1220175.1220186} {Extracting parallel
  sub-sentential fragments from non-parallel corpora}.
\newblock In \emph{Proceedings of the 21st International Conference on
  Computational Linguistics and 44th Annual Meeting of the Association for
  Computational Linguistics}, pages 81--88, Sydney, Australia. Association for
  Computational Linguistics.

\bibitem[{Pennington et~al.(2014)Pennington, Socher, and
  Manning}]{pennington-etal-2014-glove}
Jeffrey Pennington, Richard Socher, and Christopher Manning. 2014.
\newblock \href {https://doi.org/10.3115/v1/D14-1162} {{G}love: Global vectors
  for word representation}.
\newblock In \emph{Proceedings of the 2014 Conference on Empirical Methods in
  Natural Language Processing ({EMNLP})}, pages 1532--1543, Doha, Qatar.
  Association for Computational Linguistics.

\bibitem[{Peters et~al.(2018)Peters, Neumann, Iyyer, Gardner, Clark, Lee, and
  Zettlemoyer}]{peters-etal-2018-deep}
Matthew Peters, Mark Neumann, Mohit Iyyer, Matt Gardner, Christopher Clark,
  Kenton Lee, and Luke Zettlemoyer. 2018.
\newblock \href {https://doi.org/10.18653/v1/N18-1202} {Deep contextualized
  word representations}.
\newblock In \emph{Proceedings of the 2018 Conference of the North {A}merican
  Chapter of the Association for Computational Linguistics: Human Language
  Technologies, Volume 1 (Long Papers)}, pages 2227--2237, New Orleans,
  Louisiana. Association for Computational Linguistics.

\bibitem[{Pires et~al.(2019)Pires, Schlinger, and
  Garrette}]{pires-etal-2019-multilingual}
Telmo Pires, Eva Schlinger, and Dan Garrette. 2019.
\newblock \href {https://doi.org/10.18653/v1/P19-1493} {How multilingual is
  multilingual {BERT}?}
\newblock In \emph{Proceedings of the 57th Annual Meeting of the Association
  for Computational Linguistics}, pages 4996--5001, Florence, Italy.
  Association for Computational Linguistics.

\bibitem[{Pourdamghani et~al.(2019)Pourdamghani, Aldarrab, Ghazvininejad,
  Knight, and May}]{pourdamghani-etal-2019-translating}
Nima Pourdamghani, Nada Aldarrab, Marjan Ghazvininejad, Kevin Knight, and
  Jonathan May. 2019.
\newblock \href {https://doi.org/10.18653/v1/P19-1293} {Translating
  translationese: A two-step approach to unsupervised machine translation}.
\newblock In \emph{Proceedings of the 57th Annual Meeting of the Association
  for Computational Linguistics}, pages 3057--3062, Florence, Italy.
  Association for Computational Linguistics.

\bibitem[{Ramesh and
  Sankaranarayanan(2018)}]{ramesh-sankaranarayanan-2018-neural}
Sree~Harsha Ramesh and Krishna~Prasad Sankaranarayanan. 2018.
\newblock \href {https://doi.org/10.18653/v1/N18-4016} {Neural machine
  translation for low resource languages using bilingual lexicon induced from
  comparable corpora}.
\newblock In \emph{Proceedings of the 2018 Conference of the North {A}merican
  Chapter of the Association for Computational Linguistics: Student Research
  Workshop}, pages 112--119, New Orleans, Louisiana, USA. Association for
  Computational Linguistics.

\bibitem[{Reimers and Gurevych(2019)}]{reimers-gurevych-2019-sentence}
Nils Reimers and Iryna Gurevych. 2019.
\newblock \href {https://doi.org/10.18653/v1/D19-1410} {{Sentence-{BERT}:
  Sentence Embeddings using {S}iamese {BERT}-Networks}}.
\newblock In \emph{Proceedings of the 2019 Conference on Empirical Methods in
  Natural Language Processing and the 9th International Joint Conference on
  Natural Language Processing (EMNLP-IJCNLP)}, pages 3982--3992, Hong Kong,
  China. Association for Computational Linguistics.

\bibitem[{Reimers and Gurevych(2020)}]{reimers-gurevych-2020-making}
Nils Reimers and Iryna Gurevych. 2020.
\newblock \href {https://doi.org/10.18653/v1/2020.emnlp-main.365} {{Making
  Monolingual Sentence Embeddings Multilingual using Knowledge Distillation}}.
\newblock In \emph{Proceedings of the 2020 Conference on Empirical Methods in
  Natural Language Processing (EMNLP)}, pages 4512--4525, Online. Association
  for Computational Linguistics.

\bibitem[{Resnik and Smith(2003)}]{resnik-smith-2003-web}
Philip Resnik and Noah~A. Smith. 2003.
\newblock \href {https://doi.org/10.1162/089120103322711578} {The web as a
  parallel corpus}.
\newblock \emph{Computational Linguistics}, 29(3):349--380.

\bibitem[{Schuster et~al.(2019)Schuster, Ram, Barzilay, and
  Globerson}]{schuster-etal-2019-cross}
Tal Schuster, Ori Ram, Regina Barzilay, and Amir Globerson. 2019.
\newblock \href {https://doi.org/10.18653/v1/N19-1162} {Cross-lingual alignment
  of contextual word embeddings, with applications to zero-shot dependency
  parsing}.
\newblock In \emph{Proceedings of the 2019 Conference of the North {A}merican
  Chapter of the Association for Computational Linguistics: Human Language
  Technologies, Volume 1 (Long and Short Papers)}, pages 1599--1613,
  Minneapolis, Minnesota. Association for Computational Linguistics.

\bibitem[{{Schwartz} et~al.(2019){Schwartz}, {Dodge}, {Smith}, and
  {Etzioni}}]{green-ai}
Roy {Schwartz}, Jesse {Dodge}, Noah~A. {Smith}, and Oren {Etzioni}. 2019.
\newblock \href {http://arxiv.org/abs/1907.10597} {{Green AI}}.
\newblock \emph{arXiv e-prints}, page arXiv:1907.10597.

\bibitem[{Schwenk(2018)}]{schwenk-2018-filtering}
Holger Schwenk. 2018.
\newblock \href {https://doi.org/10.18653/v1/P18-2037} {Filtering and mining
  parallel data in a joint multilingual space}.
\newblock In \emph{Proceedings of the 56th Annual Meeting of the Association
  for Computational Linguistics (Volume 2: Short Papers)}, pages 228--234,
  Melbourne, Australia. Association for Computational Linguistics.

\bibitem[{{Schwenk} et~al.(2019{\natexlab{a}}){Schwenk}, {Chaudhary}, {Sun},
  {Gong}, and {Guzm{\'a}n}}]{wikimatrix}
Holger {Schwenk}, Vishrav {Chaudhary}, Shuo {Sun}, Hongyu {Gong}, and Francisco
  {Guzm{\'a}n}. 2019{\natexlab{a}}.
\newblock \href {http://arxiv.org/abs/1907.05791} {{WikiMatrix: Mining 135M
  Parallel Sentences in 1620 Language Pairs from Wikipedia}}.
\newblock \emph{arXiv e-prints}, page arXiv:1907.05791.

\bibitem[{Schwenk and Douze(2017)}]{schwenk-douze-2017-learning}
Holger Schwenk and Matthijs Douze. 2017.
\newblock \href {https://doi.org/10.18653/v1/W17-2619} {Learning joint
  multilingual sentence representations with neural machine translation}.
\newblock In \emph{Proceedings of the 2nd Workshop on Representation Learning
  for {NLP}}, pages 157--167, Vancouver, Canada. Association for Computational
  Linguistics.

\bibitem[{{Schwenk} et~al.(2019{\natexlab{b}}){Schwenk}, {Wenzek}, {Edunov},
  {Grave}, and {Joulin}}]{ccmatrix}
Holger {Schwenk}, Guillaume {Wenzek}, Sergey {Edunov}, Edouard {Grave}, and
  Armand {Joulin}. 2019{\natexlab{b}}.
\newblock \href {http://arxiv.org/abs/1911.04944} {{CCMatrix: Mining Billions
  of High-Quality Parallel Sentences on the WEB}}.
\newblock \emph{arXiv e-prints}, page arXiv:1911.04944.

\bibitem[{Sennrich et~al.(2016)Sennrich, Haddow, and
  Birch}]{sennrich-etal-2016-neural}
Rico Sennrich, Barry Haddow, and Alexandra Birch. 2016.
\newblock \href {https://doi.org/10.18653/v1/P16-1162} {Neural machine
  translation of rare words with subword units}.
\newblock In \emph{Proceedings of the 54th Annual Meeting of the Association
  for Computational Linguistics (Volume 1: Long Papers)}, pages 1715--1725,
  Berlin, Germany. Association for Computational Linguistics.

\bibitem[{{Song} et~al.(2019){Song}, {Tan}, {Qin}, {Lu}, and
  {Liu}}]{song-2019-mass2}
Kaitao {Song}, Xu~{Tan}, Tao {Qin}, Jianfeng {Lu}, and Tie-Yan {Liu}. 2019.
\newblock \href {http://arxiv.org/abs/1905.02450} {{MASS: Masked Sequence to
  Sequence Pre-training for Language Generation}}.
\newblock \emph{arXiv e-prints}, page arXiv:1905.02450.

\bibitem[{Strubell et~al.(2019)Strubell, Ganesh, and
  McCallum}]{strubell-etal-2019-energy}
Emma Strubell, Ananya Ganesh, and Andrew McCallum. 2019.
\newblock \href {https://doi.org/10.18653/v1/P19-1355} {Energy and policy
  considerations for deep learning in {NLP}}.
\newblock In \emph{Proceedings of the 57th Annual Meeting of the Association
  for Computational Linguistics}, pages 3645--3650, Florence, Italy.
  Association for Computational Linguistics.

\bibitem[{Subramanian et~al.(2018)Subramanian, Trischler, Bengio, and
  Pal}]{subramanian2018learning}
Sandeep Subramanian, Adam Trischler, Yoshua Bengio, and Christopher~J Pal.
  2018.
\newblock \href {https://openreview.net/forum?id=B18WgG-CZ} {{Learning General
  Purpose Distributed Sentence Representations via Large Scale Multi-task
  Learning}}.
\newblock In \emph{International Conference on Learning Representations}.

\bibitem[{{Tiedemann}(2020)}]{tiedemann-2020-tatoeba}
J{\"o}rg {Tiedemann}. 2020.
\newblock \href {http://arxiv.org/abs/2010.06354} {{The Tatoeba Translation
  Challenge -- Realistic Data Sets for Low Resource and Multilingual MT}}.
\newblock \emph{arXiv e-prints}, page arXiv:2010.06354.

\bibitem[{Tiedemann and Thottingal(2020)}]{tiedemann-thottingal-2020-opus}
J{\"o}rg Tiedemann and Santhosh Thottingal. 2020.
\newblock \href {https://www.aclweb.org/anthology/2020.eamt-1.61} {{{OPUS}-{MT}
  {--} Building open translation services for the World}}.
\newblock In \emph{Proceedings of the 22nd Annual Conference of the European
  Association for Machine Translation}, pages 479--480, Lisboa, Portugal.
  European Association for Machine Translation.

\bibitem[{{Tran} et~al.(2020){Tran}, {Tang}, {Li}, and {Gu}}]{tran-criss-2020}
Chau {Tran}, Yuqing {Tang}, Xian {Li}, and Jiatao {Gu}. 2020.
\newblock \href {http://arxiv.org/abs/2006.09526} {{Cross-lingual Retrieval for
  Iterative Self-Supervised Training}}.
\newblock \emph{arXiv e-prints}, page arXiv:2006.09526.

\bibitem[{Vuli{\'c} et~al.(2020)Vuli{\'c}, Ruder, and
  S{\o}gaard}]{vulic-etal-2020-good}
Ivan Vuli{\'c}, Sebastian Ruder, and Anders S{\o}gaard. 2020.
\newblock \href {https://doi.org/10.18653/v1/2020.emnlp-main.257} {{Are All
  Good Word Vector Spaces Isomorphic?}}
\newblock In \emph{Proceedings of the 2020 Conference on Empirical Methods in
  Natural Language Processing (EMNLP)}, pages 3178--3192, Online. Association
  for Computational Linguistics.

\bibitem[{Wang et~al.(2020)Wang, Xie, Xu, Yang, Neubig, and
  Carbonell}]{Wang*2020Cross-lingual}
Zirui Wang, Jiateng Xie, Ruochen Xu, Yiming Yang, Graham Neubig, and Jaime~G.
  Carbonell. 2020.
\newblock \href {https://openreview.net/forum?id=S1l-C0NtwS} {{Cross-lingual
  Alignment vs Joint Training: A Comparative Study and A Simple Unified
  Framework}}.
\newblock In \emph{International Conference on Learning Representations}.

\bibitem[{Wu et~al.(2019)Wu, Zhu, He, Gao, Qin, Lai, and
  Liu}]{wu-etal-2019-machine}
Lijun Wu, Jinhua Zhu, Di~He, Fei Gao, Tao Qin, Jianhuang Lai, and Tie-Yan Liu.
  2019.
\newblock \href {https://doi.org/10.18653/v1/D19-1446} {{Machine Translation
  With Weakly Paired Documents}}.
\newblock In \emph{Proceedings of the 2019 Conference on Empirical Methods in
  Natural Language Processing and the 9th International Joint Conference on
  Natural Language Processing (EMNLP-IJCNLP)}, pages 4375--4384, Hong Kong,
  China. Association for Computational Linguistics.

\bibitem[{Wu and Dredze(2019)}]{wu-dredze-2019-beto}
Shijie Wu and Mark Dredze. 2019.
\newblock \href {https://doi.org/10.18653/v1/D19-1077} {Beto, bentz, becas: The
  surprising cross-lingual effectiveness of {BERT}}.
\newblock In \emph{Proceedings of the 2019 Conference on Empirical Methods in
  Natural Language Processing and the 9th International Joint Conference on
  Natural Language Processing (EMNLP-IJCNLP)}, pages 833--844, Hong Kong,
  China. Association for Computational Linguistics.

\bibitem[{{Wu} et~al.(2016){Wu}, {Schuster}, {Chen}, {Le}, {Norouzi},
  {Macherey}, {Krikun}, {Cao}, {Gao}, {Macherey}, {Klingner}, {Shah},
  {Johnson}, {Liu}, {Kaiser}, {Gouws}, {Kato}, {Kudo}, {Kazawa}, {Stevens},
  {Kurian}, {Patil}, {Wang}, {Young}, {Smith}, {Riesa}, {Rudnick}, {Vinyals},
  {Corrado}, {Hughes}, and {Dean}}]{gnmt}
Yonghui {Wu}, Mike {Schuster}, Zhifeng {Chen}, Quoc~V. {Le}, Mohammad
  {Norouzi}, Wolfgang {Macherey}, Maxim {Krikun}, Yuan {Cao}, Qin {Gao}, Klaus
  {Macherey}, Jeff {Klingner}, Apurva {Shah}, Melvin {Johnson}, Xiaobing {Liu},
  {\L}ukasz {Kaiser}, Stephan {Gouws}, Yoshikiyo {Kato}, Taku {Kudo}, Hideto
  {Kazawa}, Keith {Stevens}, George {Kurian}, Nishant {Patil}, Wei {Wang},
  Cliff {Young}, Jason {Smith}, Jason {Riesa}, Alex {Rudnick}, Oriol {Vinyals},
  Greg {Corrado}, Macduff {Hughes}, and Jeffrey {Dean}. 2016.
\newblock \href {http://arxiv.org/abs/1609.08144} {{Google's Neural Machine
  Translation System: Bridging the Gap between Human and Machine Translation}}.
\newblock \emph{arXiv e-prints}, page arXiv:1609.08144.

\bibitem[{Xu et~al.(2018)Xu, Marcus, Yang, and
  Ungar}]{xu-etal-2018-unsupervised}
Hongzhi Xu, Mitchell Marcus, Charles Yang, and Lyle Ungar. 2018.
\newblock \href {https://www.aclweb.org/anthology/C18-1005} {Unsupervised
  morphology learning with statistical paradigms}.
\newblock In \emph{Proceedings of the 27th International Conference on
  Computational Linguistics}, pages 44--54, Santa Fe, New Mexico, USA.
  Association for Computational Linguistics.

\bibitem[{Zhang et~al.(2019)Zhang, Xu, Kawarabayashi, Jegelka, and
  Boyd-Graber}]{zhang-etal-2019-girls}
Mozhi Zhang, Keyulu Xu, Ken-ichi Kawarabayashi, Stefanie Jegelka, and Jordan
  Boyd-Graber. 2019.
\newblock \href {https://doi.org/10.18653/v1/P19-1307} {Are girls neko or
  sh{\=o}jo? cross-lingual alignment of non-isomorphic embeddings with
  iterative normalization}.
\newblock In \emph{Proceedings of the 57th Annual Meeting of the Association
  for Computational Linguistics}, pages 3180--3189, Florence, Italy.
  Association for Computational Linguistics.

\bibitem[{Zhao and Vogel(2002)}]{zhao-vogel-adaptive}
Bing Zhao and Stephan Vogel. 2002.
\newblock Adaptive parallel sentences mining from web bilingual news
  collection.
\newblock In \emph{Proceedings of the 2002 IEEE International Conference on
  Data Mining}, ICDM '02, page 745, USA. IEEE Computer Society.

\bibitem[{Zweigenbaum et~al.(2017)Zweigenbaum, Sharoff, and
  Rapp}]{zweigenbaum-etal-2017-overview}
Pierre Zweigenbaum, Serge Sharoff, and Reinhard Rapp. 2017.
\newblock \href {https://doi.org/10.18653/v1/W17-2512} {Overview of the second
  {BUCC} shared task: Spotting parallel sentences in comparable corpora}.
\newblock In \emph{Proceedings of the 10th Workshop on Building and Using
  Comparable Corpora}, pages 60--67, Vancouver, Canada. Association for
  Computational Linguistics.

\bibitem[{Zweigenbaum et~al.(2018)Zweigenbaum, Sharoff, and
  Rapp}]{zweigenbaum:hal-01898360}
Pierre Zweigenbaum, Serge Sharoff, and Reinhard Rapp. 2018.
\newblock \href {https://hal.archives-ouvertes.fr/hal-01898360} {{Overview of
  the Third BUCC Shared Task: Spotting Parallel Sentences in Comparable
  Corpora}}.
\newblock In \emph{{Workshop on Building and Using Comparable Corpora}},
  Miyazaki, Japan.

\end{thebibliography}
\bibliographystyle{acl_natbib}

\clearpage
\onecolumn

\appendix

\section{Appendix}

\RaggedRight

\renewcommand*{\arraystretch}{0.6}
\centering 
\footnotesize
\begin{longtable}[h]{l*{13}{c}}
\toprule
	\textbf{Procedure} & \textbf{afr} & \textbf{amh} & \textbf{ang} & \textbf{arq} & \textbf{arz} & \textbf{ast} & \textbf{awa} & \textbf{aze} & \textbf{bel} & \textbf{ben} & \textbf{ber} & \textbf
	{bos} & \textbf{bre} \\ \hline
	Raw cosine similarity \\ (\textit{Acc=F1}) & 97.4 & 94 & 64.2 & 46.2 & 78.4 & 90.6 & 73.2 & 96.1 & 96.2 & 91.3 & 10.4 & 96.2 & 17.3 \\ \\
	Margin scoring, \textit{intersection},  \\ no threshold (\textit{F1}) & 98.7 & 94.2 & \textbf{73.4} & \textbf{57.2} & \textbf{84.6} & \textbf{94.3} & \textbf{83.4} & 97.4 & 97.5 & 92.4 & \textbf{14.2} & 96.6 & \textbf{21.5} \\
	\textit{Precision} & \textit{99.9} & \textit{96.9} & \textit{88.4} & \textit{80.0} & \textit{93.6} & \textit{98.3} & \textit{95.5} & \textit{99.3} & \textit{99.1} & \textit{96.6} & \textit{30.9} & \textit{98.0} & \textit{38.5} \\
	\textit{Recall} & \textit{97.6} & \textit{91.7} & \textit{62.7} & \textit{44.5} & \textit{77.1} & \textit{90.6} & \textit{74.0} & \textit{95.6} & \textit{95.9} & \textit{88.5} & \textit{9.2} & \textit{95.2} & \textit{14.9} \\ \\
	Margin scoring, \textit{intersection}, \\ threshold = 1.06 (\textit{F1}) & 98.2 & 94.5 & 72.9 & 56.0 & 84.0 & 94.2 & 80.5 & 97.2 & 97.3 & 91.8 & 13.4 & 96.4 & 21.3 \\
	\textit{Precision} & \textit{100} & \textit{97.5} & \textit{90.1} & \textit{85.0} & \textit{95.7} & \textit{99.1} & \textit{97.0} & \textit{99.3} & \textit{99.1} & \textit{96.9} & \textit{44.4} & \textit{98.0} & \textit{54.1} \\
	\textit{Recall} & \textit{96.5} & \textit{91.7} & \textit{61.2} & \textit{41.7} & \textit{74.8} & \textit{89.8} & \textit{68.8} & \textit{95.3} & \textit{95.6} & \textit{87.3} & \textit{7.9} & \textit{94.9} & \textit{13.3} \\ \\
	Margin scoring, \textit{intersection}, \\ threshold = 1.20 (\textit{F1}) & 89.5 & 82.5 & 59.1 & 43.6 & 76.9 & 92.4 & 57.2 & 89.6 & 94.8 &  78.6 & 11.8 & 90.5 & 13.4 \\
	\textit{Precision} & \textit{100} & \textit{100} & \textit{96.6} & \textit{97.3} & \textit{98.1} & \textit{99.1} & \textit{98.9} & \textit{99.8} & \textit{99.5} & \textit{99.1} & \textit{90.0} & \textit{99.0} & \textit{92.3} \\
	\textit{Recall} & \textit{81.0} & \textit{70.2} & \textit{42.5} & \textit{28.1} & \textit{63.3} & \textit{86.6} & \textit{40.3} & \textit{81.4} & \textit{90.5} & \textit{65.1} & \textit{6.3} & \textit{83.3} & \textit{7.2} \\ \\
	Margin scoring, \textit{intersection}, \\ en-xx (\textit{F1}) & 98.4 & 93.2 & * & * & * & * & * & 96.7 & 97.6 & 91.8 & * & 96.3 & * \\
	\textit{Precision} & \textit{99.6} & \textit{96.8} & * & * & * & * & * & \textit{98.6} & \textit{99.1} & \textit{96.5} & * & \textit{98.2} & * \\
	\textit{Recall} & \textit{97.3} & \textit{89.9} & * & * & * & * & * & \textit{94.9} & \textit{96.1} & \textit{87.6} & * & \textit{94.4} & * \\ \\
	Margin scoring, \textit{intersection}, \\ xx-en (\textit{F1}) & \textbf{99.0} & \textbf{95.7} & * & * & * & * & * & \textbf{97.6} & 97.6 & 92.0 & * & 97.3 & * \\
	\textit{Precision} & \textit{99.8} & \textit{98.1} & * & * & * & * & * & \textit{99.0} & \textit{99.1} & \textit{96.3} & * & \textit{98.8} & * \\
	\textit{Recall} & \textit{98.2} & \textit{93.5} & * & * & * & * & * & \textit{96.3} & \textit{96.1} & \textit{88.0} & * & \textit{95.8} & * \\ \\
	Margin scoring, \textit{intersection}, \\ strict intersection (\textit{F1}) & 98.1 & 93.7 & * & * & * & * & * & 96.2 & 96.9 & 89.8 & * & 96.0 & * \\
	\textit{Precision} & \textit{100} & \textit{100} & * & * & * & * & * & \textit{99.8} & \textit{99.8} & \textit{99.3} & * & \textit{100} & * \\
	\textit{Recall} & \textit{96.2} & \textit{88.1} & * & * & * & * & * & \textit{92.8} & \textit{94.2} & \textit{82.0} & * & \textit{92.4} & * \\ \\
	Margin scoring, \textit{intersection}, \\ pairwise intersection (\textit{F1}) & 98.9 & 95.4 & * & * & * & * & * & 97.5 & \textbf{97.9} & \textbf{93.0} & * & 97.1 & * \\ 
	\textit{Precision} & \textit{99.9} & \textit{98.7} & * & * & * & * & * & \textit{99.3} & \textit{99.6} & \textit{97.9} & * & \textit{98.8} & * \\
	\textit{Recall} & \textit{97.9} & \textit{92.3} & * & * & * & * & * & \textit{95.9} & \textit{96.2} & \textit{88.6} & * & \textit{95.5} & * \\
	\hline
		\hline
	\textbf{Procedure} & \textbf{cbk} & \textbf{ceb} & \textbf{cha} & \textbf{cor} & \textbf{csb} & \textbf{cym} & \textbf{dsb} & \textbf{dtp} & \textbf{epo} & \textbf{eus} & \textbf{fao} & \textbf{fry} & \textbf{gla} \\ \hline
Raw cosine similarity \\ (\textit{Acc=F1}) & 82.5 & 70.9 & 39.8 & 12.8 & 56.1 & 93.6 & 69.3 & 13.3 & 98.4 & 95.8 & 90.6 & 89.9 & 88.8 \\ \\
	Margin scoring, \textit{intersection},  \\ no threshold (\textit{F1}) & \textbf{89.5} & 79.3 & \textbf{49.3} & \textbf{18.8} & \textbf{69.5} & 96.2 & \textbf{80.7} & \textbf{18.8} & \textbf{99.0} & 96.8 & \textbf{94.9} & 93.7 & 91.9 \\
	\textit{Precision} & \textit{96.7} & \textit{91.1} & \textit{65.9} & \textit{45.2} & \textit{86.5} & \textit{98.9} & \textit{94.7} & \textit{37.5} & \textit{99.7} & \textit{98.4} & \textit{98.0} & \textit{96.9} & \textit{97.1} \\
	\textit{Recall} & \textit{83.2} & \textit{70.2} & \textit{39.4} & \textit{11.9} & \textit{58.1} & \textit{93.6} & \textit{70.4} & \textit{12.5} & \textit{98.4} & \textit{95.2} & \textit{92.0} & \textit{90.8} & \textit{87.3} \\ \\
	Margin scoring, \textit{intersection}, \\ threshold = 1.06 (\textit{F1}) & 87.1 & 78.5 & 47.8 & 16.2 & 68.0 & 95.6 & 79.1 & 18.5 & 99.0 & 96.4 & 93.4 & 93.1 & 91.2 \\
	\textit{Precision} & \textit{97.8} & \textit{93.3} & \textit{75.0} & \textit{64.1} & \textit{90.2} & \textit{99.1} & \textit{95.6} & \textit{56.1} & \textit{99.9} & \textit{98.5} & \textit{98.7} & \textit{97.5} & \textit{97.3} \\
	\textit{Recall} & \textit{78.6} & \textit{67.7} & \textit{35.0} & \textit{9.3} & \textit{54.5} & \textit{92.3} & \textit{67.4} & \textit{11.1} & \textit{98.2} & \textit{94.4} & \textit{88.5} & \textit{89.0} & \textit{85.8} \\ \\
	Margin scoring, \textit{intersection}, \\ threshold = 1.20 (\textit{F1}) & 71.5 & 67.4 & 44.3 & 9.0 & 54.2 & 86.0 & 93.4 & 15.2 & 97.9 & 92.6 & 84.5 & 89.5 & 80.3 \\
	\textit{Precision} & \textit{99.6} & \textit{98.7} & \textit{85.4} & \textit{100} & \textit{95..0} & \textit{99.3} & \textit{99.6} & \textit{87.4} & \textit{99.9} & \textit{99.2} & \textit{99.0} & \textit{99.3} & \textit{98.9} \\
	\textit{Recall} & \textit{55.7} & \textit{51.2} & \textit{29.9} & \textit{4.7} & \textit{37.9} & \textit{75.8} & \textit{46.6} & \textit{8.3} & \textit{96.0} & \textit{86.8} & \textit{73.7} & \textit{81.5} & \textit{67.6} \\ \\
	Margin scoring, \textit{intersection}, \\ en-xx (\textit{F1}) & * & 78.6 & * & 15.0 & * & 96.3 & 76.2 & * & 98.5 & 96.4 & * & \textbf{96.4} & 92.6 \\
	\textit{Precision} & * & \textit{90.6} & * & \textit{36.0} & * & \textit{98.9} & \textit{95.0} & * & \textit{99.5} & \textit{98.6} & * & \textit{98.8} & \textit{97.1} \\
	\textit{Recall} & * & \textit{69.3} & * & \textit{9.5} & * & \textit{93.9} & \textit{63.7} & * & \textit{97.6} & \textit{94.3} & * & \textit{94.2} & \textit{88.4} \\ \\
	Margin scoring, \textit{intersection}, \\ xx-en (\textit{F1}) & * & \textbf{86.1} & * & 17.3 & * & \textbf{97.3} & 67.3 & * & 98.9 & \textbf{97.6} & * & 95.6 & \textbf{93.9} \\
	\textit{Precision} & * & \textit{94.2} & * & \textit{41.8} & * & \textit{98.9} & \textit{85.5} & * & \textit{99.6} & \textit{98.8} & * & \textit{97.6} & \textit{97.5} \\
	\textit{Recall} & * & \textit{79.2} & * & \textit{10.9} & * & \textit{95.7} & \textit{55.5} & * & \textit{98.3} & \textit{96.4} & * & \textit{93.6} & \textit{90.6} \\ \\
	Margin scoring, \textit{intersection}, \\ strict intersection (\textit{F1}) & * & 77.3 & * & 13.0 & * & 95.2 & 63.0 & * & 98.5 & 96.2 & * & 93.9 & 89.9 \\
	\textit{Precision} & * & \textit{99.2} & * & \textit{68.6} & * & \textit{100} & \textit{99.1} & * & \textit{100} & \textit{99.5} & * & \textit{98.7} & \textit{99.3} \\
	\textit{Recall} & * & \textit{63.3} & * & \textit{7.2} & * & \textit{90.8} & \textit{46.1} & * & \textit{97.1} & \textit{93.1} & * & \textit{89.6} & \textit{82.1} \\ \\
	Margin scoring, \textit{intersection}, \\ pairwise intersection (\textit{F1}) & * & 81.8 & * & 18.7 & * & 96.7 & 79.4 & * & 98.8 & 96.8 & * & 95.8 & 93.5 \\ 
	\textit{Precision} & * & \textit{96.0} & * & \textit{47.9} & * & \textit{99.1} & \textit{97.3} & * & \textit{99.6} & \textit{98.6} & * & \textit{98.8} & \textit{98.0} \\
	\textit{Recall} & * & \textit{71.3} & * & \textit{11.6} & * & \textit{94.4} & \textit{67.0} & * & \textit{98.1} & \textit{95.2} & * & \textit{93.1} & \textit{89.4} \\
	\hline
		\textbf{Procedure} & \textbf{gle} & \textbf{gsw} & \textbf{hsb} & \textbf{ido} & \textbf{ile} & \textbf{ina} & \textbf{isl} & \textbf{jav} & \textbf{kab} & \textbf{kaz} & \textbf{khm} & \textbf{kur} & \textbf{kzj} \\ \hline
Raw cosine similarity \\ (\textit{Acc=F1}) & 95.0
& 52.1
& 71.2
& 90.9
& 87.1
& 95.8
& 96.2
& 84.4
& 6.2
& 90.5
& 83.2
& 87.1
& 14.2 \\ \\
	Margin scoring, \textit{intersection},  \\ no threshold (\textit{F1}) & 96.6 & 62.0 & 81.6 & \textbf{95.1} & \textbf{93.0} & \textbf{97.4} & \textbf{97.9} & \textbf{92.2} & \textbf{7.7} & 92.6 & 86.8  & 92.1 & \textbf{20.8} \\
	\textit{Precision} & \textit{98.7} & \textit{85.1} & \textit{94.6} & \textit{98.7} & \textit{98.4} & \textit{99.0} & \textit{99.4} & \textit{98.9} & \textit{19.4} & \textit{96.8} & \textit{93.0} & \textit{98.1} & \textit{41.3} \\
	\textit{Recall} & \textit{94.6} & \textit{48.7} & \textit{71.8} & \textit{91.7} & \textit{88.1} & \textit{95.9} & \textit{96.4} & \textit{86.3} & \textit{4.8} & \textit{88.7} & \textit{81.3} & \textit{86.8} & \textit{13.9} \\ \\
	Margin scoring, \textit{intersection}, \\ threshold = 1.06 (\textit{F1}) & 95.9 & 60.2 & 79.7 & 94.1 & 91.7 & 96.9 & 97.5 & 91.6 & 7.3 & 92.2 & 86.4 & 91.4 & 20.0 \\
	\textit{Precision} & \textit{98.9} & \textit{89.8} & \textit{94.9} & \textit{99.0} & \textit{99.0} & \textit{99.0} & \textit{99.4} & \textit{99.4} & \textit{31.3} & \textit{96.9} & \textit{94.7} & \textit{98.3} & \textit{55.2} \\
	\textit{Recall} & \textit{93.1} & \textit{45.3} & \textit{68.7} & \textit{89.7} & \textit{85.4} & \textit{95.0} & \textit{95.7} & \textit{84.9} & \textit{4.1} & \textit{87.8} & \textit{79.5} & \textit{85.4} & \textit{12.2} \\ \\
	Margin scoring, \textit{intersection}, \\ threshold = 1.20 (\textit{F1}) & 84.7
& 43.7
& 67.8
& 88.5
& 77.9
& 94.5
& 91.0
& 83.6
& 5.0
& 85.7
& 76.4
& 82.9
& 15.1 \\
	\textit{Precision} & \textit{100}
& \textit{97.1}
& \textit{99.6}
& \textit{99.9}
& \textit{99.8}
& \textit{99.4}
& \textit{99.9}
& \textit{99.3}
& \textit{78.8}
& \textit{99.1}
& \textit{98.7}
& \textit{99.7}
& \textit{94.3} \\
  \textit{Recall} & \textit{73.5}
& \textit{28.2}
& \textit{51.3}
& \textit{79.5}
& \textit{63.8}
& \textit{90.0}
& \textit{83.6}
& \textit{72.2}
& \textit{2.6}
& \textit{75.5}
& \textit{62.3}
& \textit{71.0}
& \textit{8.2} \\ \\
	Margin scoring, \textit{intersection}, \\ en-xx (\textit{F1}) & 96.9
& 58.7
& 76.6
& 80.4
& 76.4
& 96.3
& 91.9
& *
& *
& 92.6
& 87.3
& 92.0
& * \\
	\textit{Precision} & \textit{98.8}
& \textit{80.6}
& \textit{92.9}
& \textit{91.8}
& \textit{90.1}
& \textit{99.4}
& \textit{96.4}
& *
& *
& \textit{97.0}
& \textit{93.9}
& \textit{97.5}
& * \\
	\textit{Recall} & \textit{95.2}
& \textit{46.2}
& \textit{65.2}
& \textit{71.6}
& \textit{66.3}
& \textit{93.5}
& \textit{87.8}
& *
& *
& \textit{88.7}
& \textit{81.6}
& \textit{97.1}
& * \\ \\
	Margin scoring, \textit{intersection}, \\ xx-en (\textit{F1}) & 97.7
& 59.3
& 80.0
& 82.1
& 78.7
& 95.8
& 80.8
& *
& *
& \textbf{93.5}
& 87.5
& \textbf{95.6}
& * \\
	\textit{Precision} & \textit{99.0}
& \textit{83.1}
& \textit{93.1}
& \textit{95.4}
& \textit{93.0}
& \textit{98.6}
& \textit{93.8}
& *
& *
& \textit{96.8}
& \textit{93.5}
& \textit{99.2}
& * \\
	\textit{Recall} & \textit{96.4}
& \textit{46.2}
& \textit{70.2}
& \textit{72.0}
& \textit{68.2}
& \textit{93.2}
& \textit{71.0}
& *
& *
& \textit{90.4}
& \textit{82.1}
& \textit{92.2}
& * \\ \\
	Margin scoring, \textit{intersection}, \\ strict intersection (\textit{F1}) & 95.6
& 55.1
& 74.7
& 73.2
& 67.0
& 94.9
& 78.2
& *
& *
& 91.2
& 85.6
& 90.3
& * \\
	\textit{Precision} & \textit{99.6}
& \textit{91.2}
& \textit{96.1}
& \textit{100}
& \textit{99.8}
& \textit{99.7}
& \textit{99.8}
& *
& *
& \textit{99.2}
& \textit{98.2}
& \textit{99.4}
& * \\
	\textit{Recall} & \textit{92.0}
& \textit{39.3}
& \textit{61.2}
& \textit{57.7}
& \textit{50.4}
& \textit{90.6}
& \textit{64.3}
& *
& *
& \textit{84.3}
& \textit{75.9}
& \textit{82.7}
& * \\ \\
	Margin scoring, \textit{intersection}, \\ pairwise intersection (\textit{F1}) & \textbf{97.8}
& \textbf{62.3}
& \textbf{81.7}
& 91.1
& 88.4
& 97.1
& 96.6
& *
& *
& 93.1
& \textbf{87.8}
& 94.0
& * \\ 
	\textit{Precision} & \textit{99.3}
& \textit{86.4}
& \textit{94.8}
& \textit{99.5}
& \textit{99.3}
& \textit{99.1}
& \textit{99.3}
& *
& *
& \textit{97.5}
& \textit{94.9}
& \textit{99.2}
& * \\
	\textit{Recall} & \textit{73.5}
& \textit{28.2}
& \textit{51.3}
& \textit{79.5}
& \textit{63.8}
& \textit{90.0}
& \textit{83.6}
& \textit{72.2}
& \textit{2.6}
& \textit{75.5}
& \textit{62.3}
& \textit{71.0}
& \textit{8.2} \\
	\hline
				\hline
	\textbf{Procedure} & \textbf{lat} & \textbf{lfn} & \textbf{mal} & \textbf{mhr} & \textbf{nds} & \textbf{nno} & \textbf{nov} & \textbf{oci} & \textbf{orv} & \textbf{pam} & \textbf{pms} & \textbf{swg} & \textbf{swh} \\ \hline
	Raw cosine similarity \\ (\textit{Acc=F1}) & 82.0
& 71.2
& 98.9
& 19.2
& 81.2
& 95.9
& 78.2
& 69.9
& 46.8
& 13.6
& 67.0
& 65.2
& 88.6 \\ \\
	Margin scoring, \textit{intersection},  \\ no threshold (\textit{F1}) & \textbf{89.0}
& \textbf{80.7}
& \textbf{99.3}*
& 26.3
& \textbf{89.0}
& 97.5
& \textbf{85.4}
& \textbf{78.7}
& \textbf{57.4}
& \textbf{17.9}
& \textbf{78.9}
& \textbf{80.4}
& 93.2 \\
\textit{Precision} 
& \textit{96.8}
& \textit{93.4}
& \textit{99.7}
& \textit{46.0}
& \textit{96.9}
& \textit{99.4}
& \textit{93.5}
& \textit{90.6}
& \textit{78.6}
& \textit{34.6}
& \textit{92.8}
& \textit{95.1}
& \textit{97.7} \\
	\textit{Recall} & \textit{82.4}
& \textit{71.0}
& \textit{98.8}
& \textit{18.4}
& \textit{82.2}
& \textit{95.7}
& \textit{78.6}
& \textit{69.6}
& \textit{45.3}
& \textit{12.1}
& \textit{68.6}
& \textit{69.6}
& \textit{89.0} \\ \\
	Margin scoring, \textit{intersection}, \\ threshold = 1.06 (\textit{F1}) & 87.2
& 79.4
& \textbf{99.3}*
& \textbf{26.3}
& 87.6
& 97.2
& 83.0
& 77.7
& 55.9
& 17.4
& 76.3
& 77.0
& 92.5 \\
	\textit{Precision} & \textit{97.6}
& \textit{94.7}
& \textit{99.7}
& \textit{59.3}
& \textit{98.3}
& \textit{99.5}
& \textit{94.5}
& \textit{93.1}
& \textit{83.6}
& \textit{50.2}
& \textit{94.4}
& \textit{96.0}
& \textit{98.8} \\
	\textit{Recall} & \textit{78.7}
& \textit{68.4}
& \textit{98.8}
& \textit{16.9}
& \textit{79.1}
& \textit{95.1}
& \textit{73.9}
& \textit{66.6}
& \textit{42.0}
& \textit{10.5}
& \textit{64.0}
& \textit{64.3}
& \textit{86.9} \\ \\
	Margin scoring, \textit{intersection}, \\ threshold = 1.20 (\textit{F1}) & 72.6
& 68.8
& 96.4
& 18.0
& 74.8
& 92.1
& 77.3
& 65.8
& 37.0
& 11.7
& 63.0
& 72.3
& 81.8 \\
	\textit{Precision} & \textit{99.5}
& \textit{98.5}
& \textit{99.7}
& \textit{90.1}
& \textit{99.3}
& \textit{99.9}
& \textit{98.8}
& \textit{98.8}
& \textit{96.5}
& \textit{85.1}
& \textit{98.4}
& \textit{98.5}
& \textit{100} \\
	\textit{Recall} & \textit{57.2}
& \textit{52.9}
& \textit{93.3}
& \textit{10.0}
& \textit{60.0}
& \textit{85.5}
& \textit{63.4}
& \textit{49.3}
& \textit{22.9}
& \textit{6.3}
& \textit{46.3}
& \textit{57.1}
& \textit{69.2} \\ \\
	Margin scoring, \textit{intersection}, \\ en-xx (\textit{F1}) & 83.5
& *
& 98.0
& *
& 86.0
& 97.3
& *
& *
& *
& *
& *
& *
& 94.9 \\
	\textit{Precision} & \textit{95.1}
& *
& \textit{99.5}
& *
& \textit{97.5}
& \textit{99.3}
& *
& *
& *
& *
& *
& *
& \textit{98.6} \\
	\textit{Recall} & \textit{74.4}
& *
& \textit{96.5}
& *
& \textit{76.9}
& \textit{95.4}
& *
& *
& *
& *
& *
& *
& \textit{91.5} \\ \\
	Margin scoring, \textit{intersection}, \\ xx-en (\textit{F1}) & 86.1
& *
& 98.2
& *
& 83.8
& 97.7
& *
& *
& *
& *
& *
& *
& 95.3 \\
	\textit{Precision} & \textit{95.6}
& *
& \textit{99.6}
& *
& \textit{95.2}
& \textit{99.4}
& *
& *
& *
& *
& *
& *
& \textit{98.1} \\
	\textit{Recall} & \textit{78.3}
& *
& \textit{96.9}
& *
& \textit{74.9}
& \textit{96.1}
& *
& *
& *
& *
& *
& *
& \textit{92.6} \\ \\
	Margin scoring, \textit{intersection}, \\ strict intersection (\textit{F1}) & 81.7
& *
& 97.1
& *
& 80.1
& 96.6
& *
& *
& *
& *
& *
& *
& 92.1 \\
	\textit{Precision} & \textit{98.2}
& *
& \textit{100}
& *
& \textit{99.3}
& \textit{99.8}
& *
& *
& *
& *
& *
& *
& \textit{100} \\
	\textit{Recall} & \textit{69.9}
& *
& \textit{94.3}
& *
& \textit{67.2}
& \textit{93.7}
& *
& *
& *
& *
& *
& *
& \textit{85.4} \\ \\
	Margin scoring, \textit{intersection}, \\ pairwise intersection (\textit{F1}) & 88.8
& *
& 99.2
& *
& 88.4
& \textbf{97.8}
& *
& *
& *
& *
& *
& *
& \textbf{95.5} \\ 
	\textit{Precision} & \textit{97.1}
& *
& \textit{99.9}
& *
& \textit{98.3}
& \textit{99.6}
& *
& *
& *
& *
& *
& *
& \textit{99.4} \\
	\textit{Recall} & \textit{81.7}
& *
& \textit{98.5}
& *
& \textit{80.4}
& \textit{96.0}
& *
& *
& *
& *
& *
& *
& \textit{91.2} \\
	\hline
				\hline
	\textbf{Procedure} & \textbf{tam} & \textbf{tat} & \textbf{tel} & \textbf{tgl} & \textbf{tuk} & \textbf{tzl} & \textbf{uig} & \textbf{uzb} & \textbf{war} & \textbf{wuu} & \textbf{xho} & \textbf{yid} & \textbf{} \\ \hline
Raw cosine similarity \\ (\textit{Acc=F1}) & 90.7
& 87.9
& 98.3
& 97.4
& 80.0
& 63.0
& 93.7
& 86.8
& 65.3
& 90.3
& 91.9
& 91.0
& * \\ \\
	Margin scoring, \textit{intersection},  \\ no threshold (\textit{F1}) & 93.0
& 92.0
& \textbf{99.1}*
& 98.6
& 86.8
& \textbf{71.0}
& 95.4
& 91.1
& \textbf{75.8}
& \textbf{94.8}
& 94.2
& 95.2
& * \\
	\textit{Precision} & \textit{97.5}
& \textit{97.4}
& \textit{99.6}
& \textit{99.7}
& \textit{95.8}
& \textit{82.3}
& \textit{98.3}
& \textit{96.8}
& \textit{89.5}
& \textit{98.8}
& \textit{97.7}
& \textit{98.7}
& * \\
	\textit{Recall} & \textit{88.9}
& \textit{87.1}
& \textit{98.7}
& \textit{97.6}
& \textit{79.3}
& \textit{62.5}
& \textit{92.7}
& \textit{86.0}
& \textit{65.7}
& \textit{91.1}
& \textit{90.8}
& \textit{92.0}
& * \\ \\
	Margin scoring, \textit{intersection}, \\ threshold = 1.06 (\textit{F1}) & 92.8
& 91.3
& \textbf{99.1}*
& 98.4
& 87.3
& 70.9
& 95.1
& 90.7
& 73.8
& 94.0
& 94.2
& 94.3
& * \\
	\textit{Precision} & \textit{97.8}
& \textit{97.9}
& \textit{99.6}
& \textit{99.8}
& \textit{99.4}
& \textit{87.3}
& \textit{98.3}
& \textit{97.1}
& \textit{93.5}
& \textit{99.0}
& \textit{97.7}
& \textit{99.1}
& * \\
	\textit{Recall} & \textit{88.3}
& \textit{85.5}
& \textit{98.7}
& \textit{97.1}
& \textit{77.8}
& \textit{59.6}
& \textit{92.2}
& \textit{85.0}
& \textit{60.9}
& \textit{89.4}
& \textit{90.8}
& \textit{90.0}
& * \\ \\
	Margin scoring, \textit{intersection}, \\ threshold = 1.20 (\textit{F1}) & 88.9
& 83.9
& 97.1
& 93.3
& 58.8
& 56.0
& 91.5
& 85.8
& 57.6
& 86.6
& 87.6
& 87.6
& * \\
	\textit{Precision} & \textit{98.8}
& \textit{98.9}
& \textit{100}
& \textit{100}
& \textit{98.8}
& \textit{91.3}
& \textit{99.6}
& \textit{99.4}
& \textit{99.8}
& \textit{99.5}
& \textit{97.4}
& \textit{99.5}
& * \\
	\textit{Recall} & \textit{80.8}
& \textit{72.8}
& \textit{94.4}
& \textit{87.5}
& \textit{41.9}
& \textit{40.4}
& \textit{84.6}
& \textit{75.5}
& \textit{40.5}
& \textit{76.7}
& \textit{79.6}
& \textit{78.2}
& * \\ \\
	Margin scoring, \textit{intersection}, \\ en-xx (\textit{F1}) & 93.0
& 89.8
& 98.5
& 97.5
& 85.9
& *
& 94.8
& 93.5
& *
& *
& 92.9
& 93.6
& * \\
	\textit{Precision} & \textit{98.2}
& \textit{95.4}
& \textit{99.1}
& \textit{99.2}
& \textit{95.8}
& *
& \textit{98.2}
& \textit{98.7}
& *
& *
& \textit{98.4}
& \textit{98.2}
& * \\
	\textit{Recall} & \textit{88.3}
& \textit{84.8}
& \textit{97.9}
& \textit{95.8}
& \textit{77.8}
& *
& \textit{91.6}
& \textit{88.8}
& *
& *
& \textit{88.0}
& \textit{89.5}
& * \\ \\
	Margin scoring, \textit{intersection}, \\ xx-en (\textit{F1}) & \textbf{93.7}
& \textbf{93.9}
& 97.6
& \textbf{99.4}
& \textbf{97.0}
& *
& \textbf{95.5}
& \textbf{95.2}
& *
& *
& \textbf{97.2}
& \textbf{97.2}
& * \\
	\textit{Precision} & \textit{97.5}
& \textit{97.7}
& \textit{99.1}
& \textit{99.9}
& \textit{99.5}
& *
& \textit{98.6}
& \textit{97.8}
& *
& *
& \textit{97.9}
& \textit{98.8}
& * \\
	\textit{Recall} & \textit{90.2}
& \textit{90.4}
& \textit{96.2}
& \textit{98.9}
& \textit{94.6}
& *
& \textit{92.5}
& \textit{92.8}
& *
& *
& \textit{96.5}
& \textit{95.8}
& * \\ \\
	Margin scoring, \textit{intersection}, \\ strict intersection (\textit{F1}) & 92.0
& 89.9
& 97.4
& 97.6
& 79.9
& *
& 93.7
& 91.2
& *
& *
& 91.3
& 92.7
& * \\
	\textit{Precision} & \textit{99.2}
& \textit{99.5}
& \textit{99.6}
& \textit{100}
& \textit{100}
& *
& \textit{99.7}
& \textit{100}
& *
& *
& \textit{98.4}
& \textit{99.6}
& * \\
	\textit{Recall} & \textit{85.7}
& \textit{81.9}
& \textit{95.3}
& \textit{95.3}
& \textit{66.5}
& *
& \textit{88.5}
& \textit{83.9}
& *
& *
& \textit{85.2}
& \textit{86.7}
& * \\ \\
	Margin scoring, \textit{intersection}, \\ pairwise intersection (\textit{F1}) & 93.7
& 92.5
& \textbf{99.1}*
& 98.8
& 94.0
& *
& 95.4
& 93.6
& *
& *
& 95.7
& 95.9 
& * \\ 
	\textit{Precision} & \textit{98.6}
& \textit{97.9}
& \textit{99.6}
& \textit{100}
& \textit{100}
& *
& \textit{98.7}
& \textit{99.2}
& *
& *
& \textit{98.5}
& \textit{99.1}
& *\\
	\textit{Recall} & \textit{89.3}
& \textit{87.6}
& \textit{98.7}
& \textit{97.6}
& \textit{88.7}
& *
& \textit{92.3}
& \textit{88.6}
& *
& *
& \textit{93.0}
& \textit{92.8}
& *\\
	\bottomrule \\

\captionsetup{width=1.08\textwidth}
\caption{Tatoeba test set results for a subset of low-resource language pairs, broken down by the mining method used. These language pairs are ones \textit{without} parallel data for the multilingual distillation process described in \citet{reimers-gurevych-2020-making} (cf. Table 10 in that paper). Note that LaBSE has training data for most of these languages. Descriptions of the various mining methods are found in Section \ref{Methods}.}

\label{table_tatoeba_all}
\end{longtable}


\begin{table}[ht!]
    \centering
    \footnotesize
    \renewcommand*{\arraystretch}{0.5}
    \begin{threeparttable}
    \begin{tabularx}{1.08\textwidth}{XXXXXX}
        \toprule
            \textbf{Procedure} & \textbf{Average gain over baseline (best results only)} & \textbf{Average gain over baseline (all results)} & \textbf{Average gain over baseline (langs with transl. support)} & \textbf{Best results by resource capacity*}
            & \textbf{Average gain over baseline (by resource capacity}) \\
            \hline \\
            
            Margin scoring, \textit{intersection}, no threshold
            & \textbf{+6.9} & \textbf{+5.2} & +3.6 & Level 0: 6 lang. Level 1: 18 lang. Level 2: 2 lang. Level 3: 2 lang.  \hspace{2em} 2†, 6‡ & Level 0: $+7.2$ Level 1: $+$\textbf{5.2} Level 2: $+1.8$ Level 3: $+3.4$ Level 4: $+1.0$ \\ \\
            
            Margin scoring, \textit{intersection}, threshold = 1.06
            & +2.8 & * & * & Level 0: 1 lang. Level 1: 1 lang. Level 2: 1 lang. & Level 0: $+6.1$ Level 1: $+4.3$ Level 2: $+1.6$ Level 3: $+2.9$ Level 4: $+0.6$ \\ \\
            
            Margin scoring, \textit{intersection}, threshold = 1.20
            & * & * & * & * & Level 0: $-3.8$ Level 1: $-4.3$ Level 2: $-6.8$ Level 3: $-5.2$ Level 4: $-3.2$ \\ \\
            
            Margin scoring, \textit{intersection}, en-xx
            & +6.5 & +2.4 & +2.4 & Level 0: 1 lang. & Level 0: $+4.7$ Level 1: $+1.1$ Level 2: $+0.5$ Level 3: $+2.4$ Level 4: $+0.6$ \\ \\
            
            Margin scoring, \textit{intersection}, xx-en
            & +5.2 & +3.3 & +3.3 & Level 0: 1 lang. Level 1: 7 lang. Level 2: 2 lang. Level 3: 7 lang. Level 4: 1 lang. & Level 0: $+3.9$ Level 1: $+2.8$ Level 2: $+0.1$ Level 3: $+$\textbf{4.3} Level 4: $+$\textbf{1.8} \\ \\
            
            Margin scoring, \textit{intersection}, strict intersection
            & * & * & * & * & Level 0: $+0.0$ Level 1: $-1.3$ Level 2: $-2.8$ Level 3: $-1.3$ Level 4: $+0.4$ \\ \\
            
            Margin scoring, \textit{intersection}, pairwise intersection
            & +4.6 & +4.0 & \textbf{+4.0} & Level 0: 2 lang. Level 1: 3 lang. Level 2: 2 lang. Level 3: 1 lang. & Level 0: $+$\textbf{7.3} Level 1: $+3.9$ Level 2: $+$\textbf{2.6} Level 3: $+4.0$ Level 4: $+1.0$ \\
        \bottomrule
    \end{tabularx}
    
    \begin{tablenotes}
    \item * Using resource categorizations found here: 
        \href{rb.gy/psmfnz}{rb.gy/psmfnz} \hspace{3em} † Extinct languages \hspace{3em} ‡ Constructed languages
    \end{tablenotes}
    
    \captionsetup{singlelinecheck=false, width=1.08\textwidth}

    \caption{\label{tatoeba-average} Average gain (F1) over the baseline for each mining method on the low-resource subset of the Tatoeba test data given in Table \ref{table_tatoeba_all}, broken down by several categories. The baseline is the F1 achieved using raw cosine similarity with LaBSE. The "best results" for a given method are those results on which that method achieved superior results compared to all other methods. "All results" refers to all languages in the Tatoeba test set. The "resource capacity" refers to the amount of resources a language has available. The languages with "transl. support" are those which we translated before mining (applicable for the last four methods).}
    
    \end{threeparttable}
\end{table}

\clearpage

\twocolumn

\begin{table}[ht!]
    \begin{threeparttable}
    \centering
    \footnotesize
        \begin{tabularx}{1.08\textwidth}{p{22em}p{10em}p{10em}p{10em}}
        \toprule
            \multirow{3}{*}{\textbf{Procedure}}
             & \multirow{1}{*}{\textbf{Languages on which best result was achieved (ISO 639-2 code)}} \\ \\[-3pt]
             & \multirow{1}{*}{\textbf{Gain over baseline}} \\ \\[-3pt]
             & \multicolumn{3}{l}{\textbf{Resource capacity*}} \\
             \hline \\

             Margin scoring, \textit{intersection}, no threshold & 
             ang $+9.2$ (1†) & arq $+11.0$ (?) & arz $+6.2$ (3) \\ &
             ast $+3.7$ (1) & awa $+10.2$ (0) & ber $+3.8$ (0) \\ & 
             bre $+4.2$ (1) & cbk $+7.0$ (1) & cha $+9.5$ (1) \\ & 
             cor $+6.0$ (1) & dsb $+11.4$ (0) & dtp $+5.5$ (?) \\ & 
             epo $+0.6$ (1‡) & fao $+4.3$ (1) & ido $+4.2$ (1‡) \\  & 
             ile $+5.9$ (1‡) & ila $+1.6$ (1‡) & isl $+1.7$ (2) \\ & 
             jav $+7.8$ (1) & kdz $+6.6$ (0) & ksb $+13.4$ (1) \\
             & lat $+7.0$ (3) & lfn $+9.5$ (?‡) & mal $+0.4$ (2) \\ & 
             nds $+7.8$ (0) & nov $+7.2$ (1‡) & occ $+8.8$ (1) \\ & 
             orv $+10.6$ (?†) & pam $+4.3$ (?) & pms $+11.9$ (1) \\ & 
             swg $+15.2$ (?) & tel $+0.8$ (1) & tzl $+8.0$ (?) \\ & 
             war $+10.5$ (0) & wuu $+4.5$ (1) \\ \\
             
             Margin scoring, \textit{intersection}, threshold = 1.06 & 
             mal $+0.4$ (2) & mhr $+7.1$ (0) & tel $+0.8$ (1)\\ \\
             
             Margin scoring, \textit{intersection}, threshold = 1.20 & * \\ \\ 
             
            Margin scoring, \textit{intersection}, en-xx &
            fry $+6.5$ (0)\\ \\ 
            
            Margin scoring, \textit{intersection}, xx-en &
            afr $+1.6$ (3) & amh $+1.7$ (2) & aze $+1.5$ (1) \\ & bos $+1$ (3) & ceb $+15.2$ (3) & cym $+3.7$ (1) \\ & eus $+1.8$ (4) & gla $+5.1$ (1) & kaz $+3.0$ (3) \\ & kur $+8.5$ (0) & tam $+3.0$ (3) & tat $+6.0$ (1) \\ & tgl $+2.0$ (3) & tuk $+17.0$ (1) & uig$+1.8$ (1) \\ 
            & uzb$+8.4$ (3) & xho $+5.3$ (2) & yid$+6.2$ (1) \\ \\
            
            Margin scoring, \textit{intersection}, strict intersection & * \\ \\
            
            Margin scoring, \textit{intersection}, pairwise intersection & 
            bel $+1.7$ (0) & ben $+1.7$ (3) & gle $+2.8$ (2) \\ & gsw $+10.2$ (?) & hsb $+10.5$ (0) & khm $+4.6$ (1) \\ & nno $+1.9$ (1) & swa $+6.9$ (2) & tel $+0.8$ (1) \\ \\
        \bottomrule     
    \end{tabularx}
    \begin{tablenotes}
    \item * Using resource categorizations found here: 
        \href{rb.gy/psmfnz}{rb.gy/psmfnz} \hspace{3em} † Extinct languages \hspace{3em} ‡ Constructed languages
    \end{tablenotes}
    
    \caption{LaBSE performances by mining method for each language in the Tatoeba test data. As in Tables \ref{table_tatoeba_all} and \ref{tatoeba-average}, the baseline here is F1 (accuracy) obtained using raw cosine similarity with LaBSE. The resource capacity scores are on a 0-5 scale, with 5 indicating highest availability of resources.}
    \label{tatoeba-table-list}
    \end{threeparttable}
\end{table}

\normalsize

\section{Mining on the BUCC '17/18 Training Data} \label{bucc}

\subsection{Secondary Rule-based Retrieval Methods}

\justifying
We also experimented with many rule-based mining procedures on top of margin-based mining with LaBSE on the BUCC '17/18 English-French training data. That is, we performed the initial mining pass described in Algorithm \ref{alg1}, and then used rule based metrics to filter these sentence pairs. The measures we tried included:
    \begin{itemize}[nosep]
        \item Length ratio
        \item Lexical overlap: Translate the source or target sentence, and then measure the BOW overlap
        \item Non-stopword lexical overlap
        \item Named entity overlap: Multiset named entity overlap using StanfordNER\footnote{\href{http://www.nltk.org/api/nltk.tag.html\#module-nltk.tag.stanford}{http://www.nltk.org/api/nltk.tag.html\#module-nltk.tag.stanford}} \citep{finkel-etal-2005-incorporating}.
        \item \vspace*{34em} Continuous constituent overlap: Using \citet{kitaev-etal-2019-multilingual}'s constituency parser\footnote{\href{https://github.com/nikitakit/self-attentive-parser}{https://github.com/nikitakit/self-attentive-parser}} to compute longest continuous overlap \citep{butz-wilson-xcs, lukins-2002}\footnote{\href{https://github.com/timlukins/pylcs}{https://github.com/timlukins/pylcs}} between constituents of French sentence and word-by-word translated English sentence \citep{choe-etal-2020-word2word}\footnote{\href{https://github.com/kakaobrain/word2word}{https://github.com/kakaobrain/word2word}}.
        \item BLEU score: Similar to \citet{BOUAMOR18.8}, computed BLEU score between English/French and translated French/English sentence. We experimented with NMT systems from \citet{gnmt} and \citet{tiedemann-thottingal-2020-opus}\footnote{\href{https://github.com/Helsinki-NLP/Opus-MT}{https://github.com/Helsinki-NLP/Opus-MT}} for translation, as well as word-by-word translation from \citet{choe-etal-2020-word2word} and \citet{lample2018word}\footnote{\href{https://github.com/facebookresearch/MUSE}{https://github.com/facebookresearch/MUSE}}.
        \item METEOR score: Similar to BLEU, as speculated on by \citet{BOUAMOR18.8}
        \item Hybrid methods: Combinations of the rule-based metrics above, in addition to thresholding (including ensemble thresholding using LaBSE and LASER).
    \end{itemize}
Unfortunately, none of these rule-based metrics were able to improve margin-based scoring in isolation in terms of F1, suggesting state-of-the-art similarity-based metrics have reached the level where they may not even be supplemented by rule-based metrics, including rule-ensembles, at least on high-resource language pairs.

\subsection{Confirming Flaws in Dataset}

We confirm some of the problems with the BUCC dataset that others have pointed out. In particular, we corroborate \citet{reimers-gurevych-2019-sentence}'s observation—which they make on the EN-DE data, and us on EN-FR—that the BUCC data contains many "false false positives"—that is, sentence pairs that are translations of each other but are not labeled as such. For instance, the following sentence pairs from the EN-FR train data are flagged as false positives: \\ \\
\textbf{En} According to ecological economist Malte Faber, ecological economics is defined by its focus on nature, justice, and time. \\
\textbf{Fr} \textit{Selon Malte Faber, l’économie écologique se définit par son intérêt pour la nature, la justice, et l’évolution au cours du temps.} \\
\textbf{En} Almost all parties have highly active student wings, and students have been elected to the Parliament. \\
\textbf{Fr} \textit{Presque tous les partis ont des branches universitaires très actives, et des étudiants ont été élus au Parlement.} \\
\textbf{En} Many researchers at the time strongly supported the natural selection theory. \\
\textbf{Fr} \textit{De nombreux chercheurs ont fortement soutenu la théorie de la sélection naturelle.} \\

Out of the first 100 sentence pairs flagged as false positives, we counted 72 that we would consider valid translations under rather strict criteria\footnote{\href{https://github.com/AlexJonesNLP/alt-bitexts/tree/main/BUCC_EN-FR_fp_fn}{https://github.com/AlexJonesNLP/alt-bitexts/tree/main/BUCC\_EN-FR\_fp\_fn}}. Extrapolating this to the rest of the false positives, we estimated the actual precision attainable using LaBSE with F1-based margin threshold optimization is around 97.5, in contrast with the 90.8 we originally recorded. While we don't repeat this procedure for false negatives, we notice that many of these so-called gold-standard translations suffer from coverage issues, which is why LaBSE and other similarity-based measures fail to catch them. Overall, we conclude that the actual F1 obtainable on the BUCC data with current methods is much closer to 100 than has been previously recorded \citep{reimers-gurevych-2020-making, artetxe-schwenk-2019-laser}, and we caution others against future leadboard-chasing on this benchmark, as we believe it may be "conquered."


\end{document}